\documentclass{article}

\PassOptionsToPackage{numbers, compress}{natbib}

\usepackage[final]{neurips_data_2021}

\usepackage[utf8]{inputenc} 
\usepackage[T1]{fontenc}    
\usepackage{hyperref}       
\usepackage{url}            
\usepackage{booktabs}       
\usepackage{amsfonts}       
\usepackage{nicefrac}       
\usepackage{microtype}      
\usepackage{xcolor}         
\newcommand{\bm}[1]{\boldsymbol{\mathbf{#1}}}
\usepackage{microtype}      
\usepackage{amsmath}
\usepackage{amsthm}
\usepackage{array}
\usepackage{caption}
\usepackage{mathtools}
\usepackage{mathrsfs}
\usepackage{wrapfig}
\usepackage{multirow}
\usepackage{bbm}
\usepackage{bm}
\usepackage{textcomp}
\usepackage{gensymb}
\usepackage{makecell}
\usepackage{listings}
\usepackage{xcolor}         
\usepackage{algorithm, algpseudocode}
\usepackage{tablefootnote}

\DeclareMathOperator*{\argmin}{arg\,min}
\DeclareMathOperator*{\argmax}{arg\,max}

\hypersetup{
	colorlinks=true,
	urlcolor=magenta,
}

\title{RobFR: Benchmarking Adversarial Robustness \\ on Face Recognition}

\author{%
  RobFR: Benchmarking Adversarial Robustness \\ on Face Recognition}
\author{Xiao Yang,
        Dingcheng Yang,
        Yinpeng Dong,
        Hang Su, Wenjian Yu, Jun Zhu \\
  Department of Computer Science \& Technology, Tsinghua University, Beijing, China\\
  \scriptsize{\texttt{\{yangxiao19, ydc19, dyp17\}@mails.tsinghua.edu.cn,}} 
  \scriptsize{\texttt{{\{suhangss, yu-wj, dcszj\}@tsinghua.edu.cn}}}
}

\begin{document}

\maketitle

\begin{abstract}
  Face recognition (FR) has recently made substantial progress and achieved high accuracy on standard benchmarks. However, it has raised security concerns in enormous FR applications because deep CNNs are unusually vulnerable to adversarial examples, and it is still lack of a comprehensive robustness evaluation before a FR model is deployed in safety-critical scenarios. To facilitate a better understanding of the adversarial vulnerability on FR, we develop an adversarial robustness evaluation library on FR named \textbf{RobFR}, which serves as a reference for evaluating the robustness of downstream tasks. Specifically, RobFR involves 15 popular naturally trained FR models, 9 models with representative defense mechanisms and 2 commercial FR API services, to perform the robustness evaluation by using various adversarial attacks as an important surrogate. The evaluations are conducted under diverse adversarial settings in terms of dodging and impersonation, $\ell_2$ and $\ell_\infty$, as well as white-box and black-box attacks. We further propose a landmark-guided cutout (LGC) attack method to improve the transferability of adversarial examples for black-box attacks by considering the special characteristics of FR. Based on large-scale evaluations, the commercial FR API services fail to exhibit acceptable  performance on robustness evaluation, and we also draw several important conclusions for understanding the adversarial robustness of FR models and providing insights for the design of robust FR models. RobFR is open-source and maintains all extendable modules, i.e., \emph{Datasets}, \emph{FR Models}, \emph{Attacks\&Defenses}, and \emph{Evaluations} at \url{https://github.com/ShawnXYang/Face-Robustness-Benchmark}, which will be continuously updated to promote future research on robust FR.
\end{abstract}

\section{Introduction}
\label{sec:introduction}
\vspace{-0.2ex}

Automatic face recognition (FR) is becoming a prevalent authentication solution in numerous biometric applications aiming to identify or verify a person from visual images.  
Recent progress in deep convolutional neural networks (CNNs) together with the use of abundant data~\cite{he2015deep,simonyan2014very,szegedy2015going} has greatly facilitated the massive development of FR algorithms~\cite{sun2014deep,taigman2014deepface}, which have demonstrated the state-of-the-art performance by incorporating the margin-based loss functions~\cite{deng2018arcface,liu2017sphereface,schroff2015facenet,wang2018cosface,wen2016discriminative} into deep CNN architectures. The excellent performance allows the FR models to be widely deployed in diverse scenarios ranging from financial payment to automated border control systems.


 Recent research in adversarial machine learning has revealed that deep CNNs are highly vulnerable to adversarial examples~\cite{goodfellow2014explaining,szegedy2013intriguing,yang2021boosting}, which are generated by adding a small amount of distortion to an image. 
FR systems based on deep CNNs inherit such vulnerability to adversarial examples. Considering that FR is generally safety-critical, there may exist a malicious party aiming at misleading the FR systems to achieve an illegal purpose.
For example, an adversary can evade being recognized or impersonate another identity by wearing an adversarial eyeglass~\cite{sharif2016accessorize,sharif2017adversarial,yang2020design}. Some commercial FR API services can be attacked by the adversarial examples in the black-box manner~\cite{dong2019efficient}.
As a result, adversarial attacks raise security issues to the FR systems in real-world task-critical applications, e.g., a facial payment system. It is therefore imperative to conduct a comprehensive and rigorous evaluation for a FR model before it is deployed. Although numerous enhanced methods~\cite{Dong2017,kurakin2016adversarial,papernot2016technical,rauber2017foolbox} have been proposed to conduct evaluation on image classification, it is still lack of a comprehensive evaluation on FR.

Compared with the evaluation of adversarial robustness on the conventional image classification task, benchmarking adversarial robustness on FR remains as a challenging task. First, FR models usually have more elaborate losses since FR requires to enhance high inter-class diversity and intra-class similarity simultaneously. 
It has witnessed various loss functions, e.g., Euclidean-distance-based losses~\cite{schroff2015facenet,sun2014deep,wen2016discriminative} and angular-margin-based losses~\cite{deng2018arcface,liu2017sphereface,wang2018cosface}. Second, as a pioneering application in the vision tasks, the architectures for FR are richer and diversified in that numerous attempts have been made to modify mainstream architectures~\cite{deng2019arcface,liu2017sphereface} for the adaptiveness in FR or develop assembled networks~\cite{hu2015face}. Therefore, when facing FR task with more diverse loss functions and network architectures, we still lack an in-depth understanding on whether there exists any correlation between the robustness and accuracy of the current FR models. 
Thus it is of significant importance to benchmark the adversarial robustness of the existing FR models under various settings. It will guide the users to choose proper models for different applications, and facilitate a better understanding of their robustness, which provides insights for the design of more robust models in facial tasks. 

In this paper, we establish a comprehensive and rigorous evaluation method of the adversarial robustness on FR, including white-box and black-box attack settings based on various representative FR models, covering diverse network architectures, training objectives, and defense mechanisms.
We consider face verification, an important sub-domain of face recognition aiming at distinguishing whether a pair of facial images belong to the same identity~\cite{goodfellow2014explaining}, since it has a standard well-defined testing protocol for the comparison between different algorithms but can also provide a good reference for other sub-tasks, e.g., face identification. 
Specifically, we incorporate $15$ popular naturally trained FR models, which cover most of the models trained by representative losses and network architectures. We also involve $9$ representative defense models including input transformation~\cite{dziugaite2016study,xu2018feature} and adversarial training~\cite{brendel2020adversarial} in the FR task for the first time. we adopt several evaluation metrics, including 1) attack success rate (Asr); 2) \emph{Asr vs. perturbation budget robustness} curves that can give a global understanding of the robustness of FR models; 3) transfer-based Asr that can directly reflect black-box evaluation results by using adversarial examples generated against a substitute model. 
Moreover, we carry out large-scale experiments under thorough threat models to fully demonstrate the performance of different FR models, including 1) dodging and impersonation attacks; 2) $\ell_2$ and $\ell_\infty$ attacks; 3) white-box and black-box attacks that can reflect the evaluation results in different levels. Thus they are recommended as the robustness evaluation metrics for future works on FR. 



Since the existing adversarial libraries (e.g., CleverHans~\cite{papernot2016technical}, Foolbox~\cite{rauber2017foolbox}, FACESEC~\cite{tong2021facesec}, ARES~\cite{dong2020benchmarking}) cannot fully support our evaluations in FR (see Appendix {\color{red} A}). In this paper, we develop a new adversarial robustness evaluation library on face recognition named \textbf{RobFR} to conduct all experiments. RobFR consists of four modules, \emph{i.e.}, Datasets, FR Models, Attacks \& Defenses, and Evaluations, which takes a easily extendable implementation due to independent interface, thus enabling more researchers to conveniently supplement new contents. In RobFR we further propose a \emph{landmark-guided cutout} (LGC) attack method in RobFR to improve the performance of black-box attacks by considering the particular characteristics of FR.
LGC can be flexibly incorporated into black-box attack methods to improve their performances, such as MIM~\cite{Dong2017}, DIM~\cite{xie2019improving}, and TIM~\cite{dong2019evading}.

By analyzing the evaluation results, we have some important findings. First, the excellent algorithms with glorious precision in FR, e.g., ArcFace~\cite{deng2019arcface} and CosFace~\cite{wang2018cosface}, fail to substantially improve the robustness.
Considering the safety-critical requirement, future FR research should significantly take into account the robustness when pursuing a high precision.
Second, model architecture is a more critical factor for boosting robustness than other factors, e.g., loss functions. Thus selecting a proper larger network structure as a backbone usually exhibits better resistance against adversarial examples, which also provides an insight of designing robust FR models for the future. Third, adversarial training (AT) is the most robust method among the various defense methods, whereas AT results in a degeneration of natural accuracy and high training cost. More discussions are provided in Sec.~\ref{sec:discussion}.





\vspace{-1.5ex}
\section{Background and Threat Models}
\vspace{-0.5ex}

Let $f(\bm{x}): \mathcal{X}\rightarrow\mathbb{R}^d$ denote a FR model that extracts a normalized feature representation in $\mathbb{R}^d$ for an input face image $\bm{x}\in\mathcal{X}\subset\mathbb{R}^n$. Face verification aims to compute the distance between the feature representations of image pair $\{\bm{x}_{1}, \bm{x}_{2}\} \subset \mathcal{X}$. We first denote their feature distance $\mathcal{D}_f(\bm{x}_{1}, \bm{x}_{2})$ as
\begin{equation}
    \mathcal{D}_f(\bm{x}_{1}, \bm{x}_{2}) = \|f(\bm{x}_1) - f(\bm{x}_2)\|_2^2.
\end{equation}
Face verification can therefore be formally formulated as
\begin{equation}\label{eq:delta}
    \mathcal{C} (\bm{x}_{1}, \bm{x}_{2}) = \mathbb{I} (\mathcal{D}_f (\bm{x}_{1}, \bm{x}_{2}) < \delta ),
\end{equation}
where $\mathbb{I}$ is the indicator function, and $\delta$ is a threshold.  When $\mathcal{C}(\bm{x}_{1}, \bm{x}_{2})$ equals to $1$, the image pair is recognized as the same identity and otherwise different identities. Notely, this definition is consistent with the commonly used cosine similarity metric, since $f$ usually outputs a normalized vector.

Given a FR model, we need to specify the threat models for robustness evaluation, including different goals, capabilities, and knowledge~\cite{carlini2019evaluating} of the adversary.

\textbf{Adversary's goals.} We focus on two types of attacks in terms dodging and impersonation. Specifically, \emph{dodging attacks} seeks to mislead a face recognition system thus have one face misidentified. In general, dodging attacks are of great interests in evading surveillance. Formally, given a pair
of face images $\bm{x}$ and $\bm{x}^{r}$ with the \textit{same} identity, the adversary
seeks to modify $\bm{x}$ to generate an adversarial image $\bm{x}^{adv}$ that cannot be recognized by the model, i.e., to make $\mathcal{C} (\bm{x}^{adv}, \bm{x}^{r}) = 0$. On the other hand, \emph{impersonation attacks} aim to generate an adversarial example that can be recognized as a specific victim’s identity, which generally harder than dodging. One attackers can leverage this method to evade the face authentication systems. Formally, given a pair
of face images $\bm{x}$ and $\bm{x}^{r}$ with two \textit{different} identities,
the adversary tries to find an adversarial image $\bm{x}^{adv}$ that is recognized as the target identity of $\bm{x}^r$, i.e., to make $\mathcal{C} (\bm{x}^{adv}, \bm{x}^{r}) = 1$.

\textbf{Adversary's capabilities.} As adversarial examples are usually assumed to be indistinguishable from the corresponding original ones from human observations~\cite{goodfellow2014explaining,szegedy2013intriguing}, the adversary can only introduce small changes to the inputs. In this paper, we study widely used $\ell_p$ additive perturbation setting, where the adversary is allowed to add a small perturbation measured by the $\ell_{\infty}$ and $\ell_2$ norms to the original input.
To achieve the adversary's goal, we optimize the feature distance between the adversarial image $\bm{x}^{adv}$ and the counterpart face image $\bm{x}^r$, meanwhile, we keep a small distance between $\bm{x}^{adv}$ and $\bm{x}$ in the input space.  
For dodging attacks, an adversarial image is generated by maximizing the distance between $\bm{x}'$ and $\bm{x}^{r}$ in the feature space as
\begin{equation}
\label{eq:constrain-pert-d}
 \bm{x}^{adv}= \argmax_{\bm{x}':\|\bm{x}'-\bm{x}\|_p\leq\epsilon}\mathcal{D}_f(\bm{x}',\bm{x}^{r}),
\end{equation}
where $\epsilon$ is a small constant. By solving the problem in Eq.~\eqref{eq:constrain-pert-d}, the FR model will misclassify them as different identities when the feature distance is larger than a predefined threshold $\delta$.
For impersonation attacks, we can formulate the problem as minimizing the distance between $\bm{x}'$ and $\bm{x}^{r}$ as 
\begin{equation}
\label{eq:constrain-pert-p}
 \bm{x}^{adv}= \argmin_{\bm{x}':\|\bm{x}'-\bm{x}\|_p\leq\epsilon}\mathcal{D}_f(\bm{x}',\bm{x}^{r}).
\end{equation}
Therefore, the feature representation of $\bm{x}^{adv}$ will resemble that of $\bm{x}^r$, such that they are recognized as the same identity by FR model.

\textbf{Adversary's knowledge.} An adversary can have different levels of knowledge of the target FR models to craft adversarial examples. Besides the white-box access to the model information, the black-box scenario is another important setting to evaluate the robustness in FR, since the black-box attack setting is more practical in real-world applications, e.g., breaking into a commercial FR API. 
Thus we consider both \emph{white-box attacks} and \emph{black-box attacks} in this paper. White-box attacks rely on detailed information of the target FR model, including architectures, parameters, and gradients of the loss w.r.t. the input. Black-box attacks can be realized based on the transferability of adversarial examples~\cite{Papernot2016}. Transfer-based black-box attacks do not rely on model information but assume the availability of a substitute model based on which the adversarial examples can be generated.

\vspace{-1ex}
\section{Benchmarking Adversarial Robustness on Face Recognition}
\vspace{-0.2ex}


\begin{figure*}[t]
\centering
\includegraphics[width=0.99\linewidth]{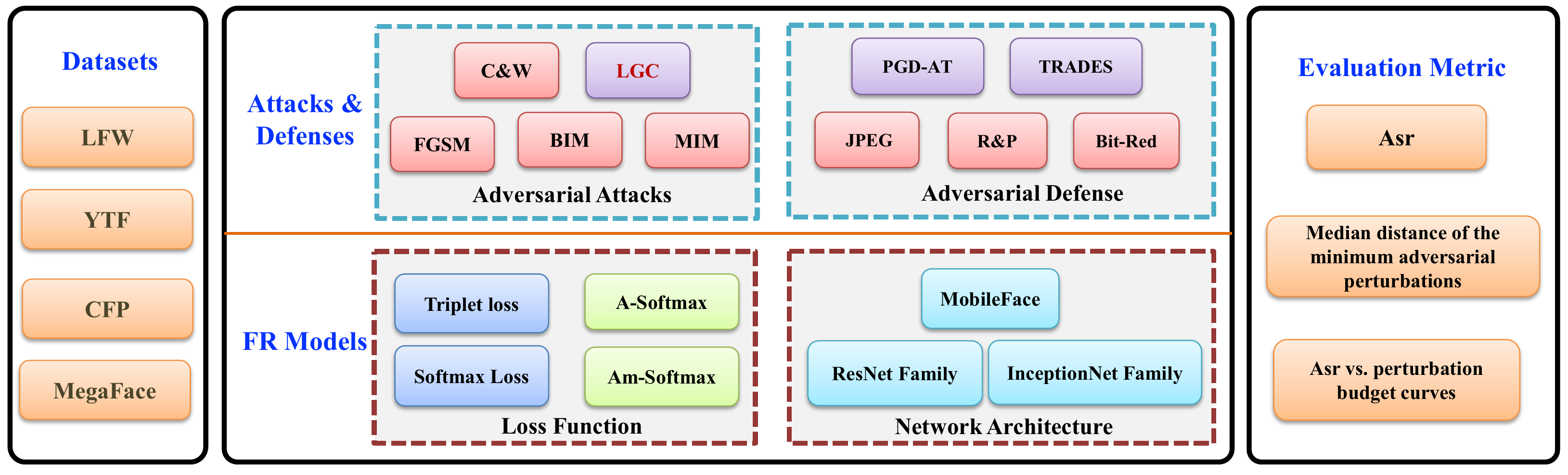}
\vspace{-1ex}
\caption{The overall framework of RobFR library. The module of \textbf{Datasets} considers current mainstream and more challenging datasets, to perform robustness evaluations. The module of \textbf{FR Models} includes $15$ representative and popular FR models, covering different \emph{model architectures} including mainstream architectures (e.g., ResNet and InceptionNet families) and light-weight networks (e.g., MobileFace), and \emph{training loss functions} including Euclidean-distance-based losses (e.g., Softmax and Triplet loss) and angular-margin-based losses (e.g., A-Softmax and AM-Softmax). The module of \textbf{Attacks $\&$ Defenses} incorporates some representative attack and defense algorithms to adapt the FR task. Attacks include \emph{white-box} ones, such as gradient-based FGSM, BIM and MIM and optimization-based C\&W, and \emph{black-box} ones, such as FGSM, BIM, MIM and \textbf{LGC} proposed by us. Defenses incorporate two general and representative categories, including  \emph{input transformation}, such as JPEG compression, R\&P and bit-depth reduction, and \emph{adversarial training}, such as PGD-AT and TRADES. The module of \textbf{Evaluation Metric} develops three metrics to evaluate adversarial robustness in FR.}
\label{fig:framework}
\vspace{-3ex}

\end{figure*}



The existing adversarial machine learning platforms, including CleverHans~\cite{papernot2016technical}, Foolbox~\cite{rauber2017foolbox}, ARES~\cite{dong2020benchmarking}, etc., focus on the robustness evaluation in image classification. 
As a comparison, RobFR implements a comprehensive robustness evaluation in FR as illustrated in Fig.~\ref{fig:framework}, which consists of four modules, \emph{i.e.}, Datasets, FR Models, Attacks \& Defenses, and Evaluations. 
Notably, our library takes an easily extendable implementation for every module due to independent interface, thus enabling more researchers to conveniently supplement new contents.

\vspace{-1ex}
\subsection{FR Models} 
\vspace{-0.3ex}

\begin{wraptable}{r}{0.52\linewidth}
    \scriptsize
    \vspace{-5ex}
    \caption{The FR models adopted for the robustness evaluation in this paper. $^{*}$ refers to public checkpoints.}
    \vspace{-1ex}
    \label{tab:models}
    \setlength{\tabcolsep}{1pt}
    \begin{tabular}{c|c|c|c|c|c}
    \toprule
    Models  & Backbone & Loss  & Para.(M) & Acc.(\%) & Thr.  \\ 
    \midrule
    FaceNet$^{*}$ & InceptionResNet & Triplet  & 27.91 & 99.2 & 1.159\\ 
    SphereFace$^{*}$ & Sphere20 & A-Softmax & 28.08 & 98.2 & 1.301 \\ 
    CosFace$^{*}$ & Sphere20 & LMCL & 22.67 &  98.7 & 1.507  \\
    ArcFace$^{*}$ & IR-SE50 & Arcface   & 43.80 & 99.5 & 1.432\\  \midrule
    MobleFace  & MobileFaceNet & LMCL & 1.20 & 99.5 & 1.578 \\
    MobileNet & MobileNet & LMCL & 3.75 & 99.4 & 1.683 \\ 
    MobileNetv2 & MobileNetv2 & LMCL & 2.90 & 99.3 & 1.547 \\ 
    ShuffleNetV1 & ShuffleNetV1 & LMCL & 1.46 & 99.5 & 1.619 \\ 
    ShuffleNetV2 & ShuffleNetV2 & LMCL & 1.83 & 99.2 & 1.552 \\  
    ResNet50 & ResNet50 & LMCL & 40.29 & 99.7 & 1.618 \\
    \midrule
    Softmax-IR & IResNet50 & Softmax & 43.57 & 99.6 & 1.315 \\ 
    SphereFace-IR & IResNet50 & A-Softmax & 43.57 & 99.6 & 1.277\\
    AM-IR &  IResNet50 & AM-Softmax & 43.57 & 99.2 & 1.083\\ 
    CosFace-IR & IResNet50 & LMCL & 43.57 & 99.7 & 1.552\\
    ArcFace-IR & IResNet50 & ArcFace & 43.57 & 99.7 & 1.445\\
    \midrule
    {\color{blue} Microsoft API} & - & - & - & 98.8 & 0.326\\
    {\color{blue} Tencent API} & - & - & - & 99.5 & 51.23\\
    \bottomrule
    
    \end{tabular}
    \vspace{-3ex}
\end{wraptable}

To effectively evaluate the robustness of FR models, we choose $15$ representative and state-of-the-art FR models, covering different model architectures and training loss functions, as shown in Tab.~\ref{tab:models}. We first include the best publicly available models FaceNet~\cite{schroff2015facenet}, SphereFace~\cite{liu2017sphereface}, CosFace~\cite{wang2018cosface}, and ArcFace~\cite{deng2019arcface} for evaluations. Besides, to measure the effects of model architectures, we train many models under different backbones with multiple sizes of weights, such as the mainstream architecture ResNet50~\cite{he2016}, and light-weight networks including MobileFace~\cite{chen2018mobilefacenets}, MobileNet~\cite{howard2017mobilenets},  MobileNetV2~\cite{sandler2018mobilenetv2}, ShuffleNetV1~\cite{zhang2018shufflenet}, and ShuffleNetV2~\cite{ma2018shufflenet}, which are trained by the advanced LMCL loss.
Moreover, to study the effects of different training loss functions, we include the models of the same evolved IResNet50~\cite{deng2019arcface} architecture in FR optimized by distance-based losses, such as Softmax, and those optimized by angular-margin-based losses, such as Sphereface,
AM-Softmax, CosFace, and ArcFace. We also involve two commercial API services (see Appendix {\color{red} E}), and the working mechanism and training data are completely unknown for us. For each model used in our evaluations, we first compute the optimal threshold\footnote{The Euclidean threshold $\delta_{d}$ can be obtained from the Cosine threshold $\delta_{c}$ after normalizing features as $\delta_{d}= 2 - 2\cdot\delta_{c}$, where $\delta_{d} \in[0, 4]$. As for APIs, we follow original threshold ranges to compute optimal values.} in Eq.~\eqref{eq:delta} of every FR model by following the standard protocol from the LFW dataset, as shown in Tab.~\ref{tab:models}. If the distance of two images that are fed into the model exceeds the threshold, we regard them as different identities, otherwise same identities. 

\vspace{-1ex}
\subsection{Datasets}
\vspace{-0.3ex}
LFW~\cite{huang2008labeled} and YTF~\cite{wolf2011face} are two of the most widely used benchmark datasets for face verification on images and videos. Thus we elaborate the testing protocols via dodging and impersonation attacks to adaptively benchmark the adversarial robustness in FR models. Besides, to further explore robustness evaluations on more challenging datasets, we also perform the evaluation on CFP-FP~\cite{Sengupta2016Frontal} and MegaFace~\cite{kemelmacher2016megaface}.
Due to numerous individuals from MegaFace,  in this study we randomly choose 6,000 pairs to implement the robustness evaluation from standard verification protocols.

\textbf{Implementation details.} We perform dodging attacks based on the pairs of images with the same identities, and impersonation attacks based on the pairs with different identities. We first use MTCNN~\cite{zhang2016joint} to detect facial area and align images for the entire images. Then, we obtain the cropped faces which are resized to $112\times112$. Note that the specific input size and pixel transformation will be executed inside each model due to the diversity of model inputs. 

\vspace{-1ex}
\subsection{Evaluation Metric} 
\vspace{-0.2ex}

Given an attack method $\mathcal{A}_{\epsilon, p}$ that generates an adversarial example $\bm{x}^{adv} = \mathcal{A}_{\epsilon, p}(\bm{x}, \bm{x}^{r})$ for an input $\bm{x}$ and a reference image $\bm{x}^r$ with perturbation budget $\epsilon$ under the $\ell_p$ norm ($\|\bm{x}^{adv} - \bm{x}\|_p \leq \epsilon$), the first evaluation metric is the attack success rate (Asr) on the FR model $\mathcal{C}$ in Eq.~\eqref{eq:delta}, defined as
\begin{equation}\small
    \mathrm{Asr}(\mathcal{C}, \mathcal{A}_{\epsilon, p}) = \frac{1}{N}\sum_{i=1}^{N}\mathbb{I}\big(\mathcal{C}(\mathcal{A}_{\epsilon, p}(\bm{x}_i, \bm{x}_{i}^{r}), \bm{x}_i^r) \neq y_i\big),
\end{equation}
where $(\bm{x}_i, \bm{x}_i^r)_{i=1}^{N}$ is the paired test set, $y_{i}$ takes $1$ if the pair belongs to same identity, and $0$ otherwise.
The second evaluation metric is the median distance of the \textit{minimum adversarial perturbations}, which is a metric to show the worst-case robustness of the models~\cite{brendel2020adversarial}. The objective of minimum adversarial perturbations can be denoted as
\begin{equation}\small
    \min \epsilon, \;\text{s.t.}\;
    \mathcal{C}(\mathcal{A}_{\epsilon, p}(\bm{x}_i, \bm{x}_{i}^{r}), \bm{x}_i^r) \neq y_i.
\end{equation}
To obtain this value for each data, we perform a binary search on $\epsilon$ to find its minimum value that fulfills the adversary's goal.


The third evaluation metric is the \emph{Asr vs. perturbation budget} curve, which can give a global understanding of the robustness of FR models~\cite{dong2020benchmarking}.
To obtain this curve, we need to calculate Asr for all values of $\epsilon$, which can be efficiently done by finding the minimum perturbations and counting the number of the adversarial examples, the $\ell_p$ norm of whose perturbations is smaller than each $\epsilon$. We use different criteria for different attacks, which will be specified in Sec.~\ref{sec:exp}.

\vspace{-1ex}
\subsection{Traditional Attack Methods}
\vspace{-0.2ex}

To solve the problem in Eq.~\eqref{eq:constrain-pert-d} or Eq.~\eqref{eq:constrain-pert-p}, many methods can be used to generate adversarial examples. In this section, we summarize the typical white-box and black-box adversarial attack methods that are adopted for robustness evaluations. We only introduce these methods for dodging attacks, since the extension to impersonation attacks is straightforward.

\textbf{Fast Gradient Sign Method (FGSM)}~\cite{goodfellow2014explaining} generates an adversarial example given a pair of images $\bm{x}$ and $\bm{x}^{r}$ with the same identity under the $\ell_{\infty}$ norm and $\ell_2$ norm as 
\begin{equation}\small
    \label{eq:fgsm}
    \bm{x}^{adv} = \bm{x} + \epsilon\cdot\mathrm{sign}(\nabla_{\bm{x}}\mathcal{D}_f(\bm{x}, \bm{x}^{r})),\quad \bm{x}^{adv} = \bm{x} + \epsilon\cdot\frac{\nabla_{\bm{x}}\mathcal{D}_f(\bm{x},\bm{x}^{r})}{\|\nabla_{\bm{x}}\mathcal{D}_f(\bm{x},\bm{x}^{r})\|_2},
\end{equation}
where $\nabla_{\bm{x}}\mathcal{D}_f$ is the gradient of the feature distance w.r.t. $\bm{x}$, and $\mathrm{sign(\cdot)}$ is the sign function to make the perturbation meet the $\ell_{\infty}$ norm bound.

\textbf{Basic Iterative Method (BIM)}~\cite{Kurakin2016} extends FGSM by iteratively taking many gradient updates as
\begin{equation}\small
\label{eq:bim}
    \bm{x}_{t+1}^{adv} = \mathrm{clip}_{\bm{x},\epsilon} \big(\bm{x}_t^{adv} + \alpha\cdot\mathrm{sign}(\nabla_{\bm{x}}\mathcal{D}_f(\bm{x}_t^{adv},\bm{x}^{r}))\big),
\end{equation}
where $\mathrm{clip}_{\bm{x},\epsilon}$ projects the adversarial example to satisfy the $\ell_{\infty}$ constrain and $\alpha$ is the step size. We only adopt BIM for evaluations since PGD~\cite{madry2017towards} and BIM result in similar attack performance. 

\textbf{Momentum Iterative Method (MIM)}~\cite{Dong2017} integrates a momentum term into BIM for improving the transferability of adversarial examples as 
\begin{equation}\small
\label{eq:mim}
\begin{gathered}
     \bm{g}_{t+1} = \mu \cdot \bm{g}_t + \frac{\nabla_{\bm{x}}\mathcal{D}_f(\bm{x}_t^{adv},\bm{x}^{r})}{\|\nabla_{\bm{x}}\mathcal{D}_f(\bm{x}_t^{adv},\bm{x}^{r})\|_1}; \quad  \bm{x}^{adv}_{t+1}=\mathrm{clip}_{\bm{x},\epsilon}(\bm{x}^{adv}_t+\alpha\cdot\mathrm{sign}(\bm{g}_{t+1})).
\end{gathered}
\end{equation}

\textbf{Carlini \& Wagner's Method (C\&W)}~\cite{carlini2017towards} is a powerful optimization-based attack method. It takes a Lagrangian form of the constrained optimization problem and adopts Adam~\cite{Kingma2014} for  optimization, which is quite effective for $\ell_2$ attacks. However, the direct extension of the C\&W method to FR is problematic since C\&W uses the loss function defined on the logits of the classification models.
In FR, there is not any logit output of the FR models. Thus we define a new attack objective function suitable for FR systems.
For dodging attacks, the optimization problem can be formulated as 
\begin{equation}\small
\label{eq:cw}
    \bm{x}^{adv} = \argmin_{\bm{x}'} \big\{\|\bm{x}'-\bm{x}\|_2^2 + c\cdot\max(\delta - \mathcal{D}_f(\bm{x}', \bm{x}^{r}), 0)\big\},
\end{equation}
where $c$ is a parameter to balance the two loss terms, whose optimal value is discovered by binary search. C\&W is not good at $\ell_{\infty}$ attacks~\cite{carlini2017towards}, so we only use it under the $\ell_2$ norm.

\textbf{Transfer-based Black-box Attacks.} Under this setting, we generate adversarial examples by using the above methods against a substitute FR model, and use them to attack the black-box models. For the studied $15$ FR models in our evaluations, we treat each one as the white-box model to generate adversarial examples and test the performance against the other models.

\vspace{-1ex}
\subsection{Landmark-Guided Cutout (LGC) Attack}
\vspace{-0.3ex}

\begin{figure}[t]
\centering
\includegraphics[width=0.9\linewidth]{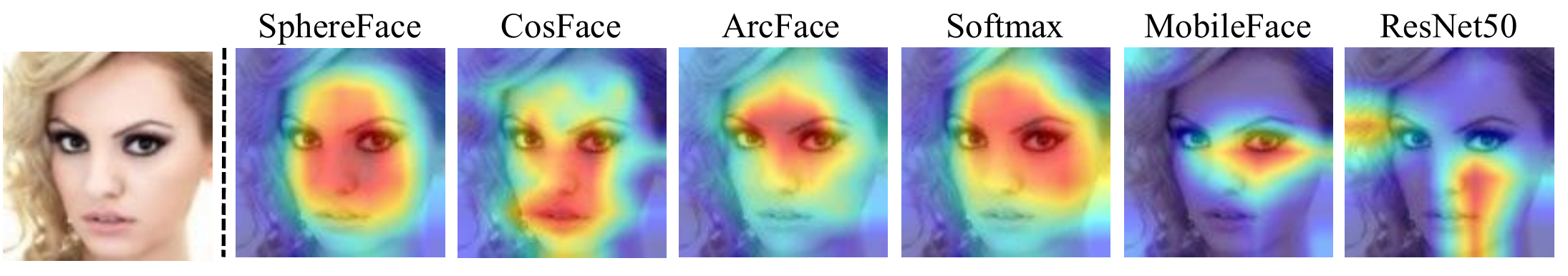}
\vspace{-2ex}
\caption{The illustration of the attention maps of the different models. We employ Grad-Gram~\cite{selvaraju2017grad} which produces the attention maps highlighting the discriminative regions.}
\label{fig:att}
\vspace{-4ex}
\end{figure}

Although the above attack methods can be used for attacking FR models, they are originally designed on the general image classification tasks. The adversarial examples can be highly associated with the discriminative local region of the white-box model for an input image, as emphasized in~\cite{dong2019evading}. The normally trained image classification models generally have similar discriminative regions, making the crafted adversarial examples have a high transferability~\cite{dong2019evading,tsipras2018robustness}. However, we find that the existing SOTA FR models have different attention maps for their predictions given the same input image, as illustrated in Fig.~\ref{fig:att}. Therefore, the crafted adversarial examples will depend on the discriminative local region of the white-box model, making it difficult to transfer to the black-box models with different discriminative regions.

To weaken the effect of different discriminative regions for the overall transferability, we further propose a \textbf{landmark-guided cutout} (LGC) attack method. The \emph{cutout} method~\cite{devries2017improved} has been proven that it can make models take image context into consideration by randomly occluding units at the input space. Therefore, selecting appropriate occlusion from key positions can reduce sensibility of discriminative regions in the process of generating adversarial examples. Many studies have also shown that there exist some important regions for FR, including the surroundings of eyes and noses~\cite{gutta2002investigation,yang2019face}, which are also consistent with the overall prominent regions shown in Fig.~\ref{fig:att}. Thus we extend \emph{Cutout} to the proposed LGC method, which occludes units from \emph{prominent} input regions of the images, making the network focus on less prominent regions and obtain more transferable adversarial examples. To achieve this goal, we apply a spatial prior by locating face landmarks by using face landmark detection~\cite{gutta2002investigation}.  Specifically, we apply a fixed-size mask to randomly sampled $m$ locations as center points from face landmarks, and place a square patch around those locations. A simple combination of MIM and landmark-guided cutout can give rise to the LGC method as
\begin{equation}\small
\begin{gathered}
\label{eq:cim}
     \bm{g}_{t+1} = \mu \cdot \bm{g}_t + \frac{\nabla_{\bm{x}}\mathcal{D}_f({M}_t\odot{\bm{x}_t^{adv}},\bm{x}^{r})}{\|\nabla_{\bm{x}}\mathcal{D}_f({M}_t\odot{\bm{x}_t^{adv}},\bm{x}^{r})\|_1}; \quad
     \bm{x}^{adv}_{t+1}=\mathrm{clip}_{\bm{x},\epsilon}(\bm{x}^{adv}_t+\alpha\cdot\mathrm{sign}(\bm{g}_{t+1})),
\end{gathered}
\end{equation}
where $M_{t} \in \{0,1\}^{d}$ is a binary mask, $\odot$ is the element-wise dot product, and $d$ is the dimension of the face image.
In the $t$-th iteration, after initializing the values of $M_{t}$ as $1$, the some randomly sampled fixed-size small square regions are set to $0$ to form $M_{t}$. The algorithm is summarized in Appendix {\color{red} C.1}. We regard the combination of LGC and MIM as \textbf{LGC attack} as default in this paper. Besides, we demonstrate that the proposed LGC can also be incorporated into other black-box attack methods to improve their performances, such as DIM~\cite{xie2019improving} and TIM~\cite{dong2019evading} in Appendix {\color{red} C.4}.

\vspace{-1ex}
\subsection{Defense Methods}
\vspace{-0.3ex}

Extensive research has concentrated on building robust models against adversarial attacks on image classification. Below we introduce two general and representative categories to adapt the FR task.

\textbf{Input Transformation.} Some defenses transform the inputs before feeding them into networks, including JPEG compression~\cite{dziugaite2016study} and bit-depth reduction~\cite{xu2018feature}, or add some randomness on the input~\cite{xie2017mitigating}. We incorporate these input transformations into the natural FR models for defenses. 

\textbf{Adversarial Training (AT).} AT is one of the most effective methods on defending adversarial attacks~\cite{brendel2020adversarial,kurakin2018competation,zhang2019theoretically}. PGD-AT~\cite{madry2018towards}, as the most popular one, formulates the AT procedure as a min-max problem. Therefore, AT methods in FR can be formulated as the two-stage framework:
\begin{equation}\small
\label{eq:at}
    \min_{\bm{\omega},\mathbf{W}}\frac{1}{n}\sum_{i=1}^{n}\max_{\bm{\eta }_{i}\in \mathcal{S}} \mathcal{L}(f(\bm{x}_{i} + \bm{\eta}_{i}), y_{i}, \mathbf{W}),
\end{equation}
where $f(\cdot)$ is the feature extractor with parameters $\bm{\omega}$, the matrix $\mathbf{W} = (W_1,...,W_C)$ is the weight matrix for the task with $C$ labels, $\mathcal{L}$ is a cross-entropy loss, and $\mathcal{S}=\{\bm{\eta}:\|\bm{\eta}\|_{\infty} \leq \epsilon\}$ is a set of allowed points around $\bm{x}$ with the perturbation $\epsilon$. Adversarial examples are crafted in the inner maximization, and model parameters are optimized by solving the outer minimization. Therefore, they are iteratively executed in training until model parameters $\bm{\omega}$ and $\mathbf{W}$ converge. PGD~\cite{madry2018towards} has been extensively applied in the inner optimization, which is denoted by taking multiple steps as 
\begin{equation}\small
    \bm{\eta}_{i}^{t+1} = \prod_{\mathcal{S}}\big(\bm{\eta}_i^{t} + \alpha\cdot\mathrm{sign}(\nabla_{\bm{x}}\mathcal{L}(f(\bm{x}+\bm{\eta}_{i}^{t}),y_{i},\mathbf{W}))\big),
\end{equation} 
where $\bm{\eta}_{i}$ is the perturbation at the t-th step, $\alpha$ is the step size and $\prod(\cdot)$ is a projection function in $\mathcal{S}$.
In this paper, we mainly consider the representative AT framework named PGD-AT~\cite{madry2018towards}, and adaptively integrate different loss functions (as detailed in Appendix {\color{red} D}) into Eq.~\eqref{eq:at} with an AT procedure. Besides, another typical framework TRADES~\cite{zhang2019theoretically} is also evaluated as a comparison.




\begin{figure*}[t]
\begin{center}
\includegraphics[width=0.99\linewidth]{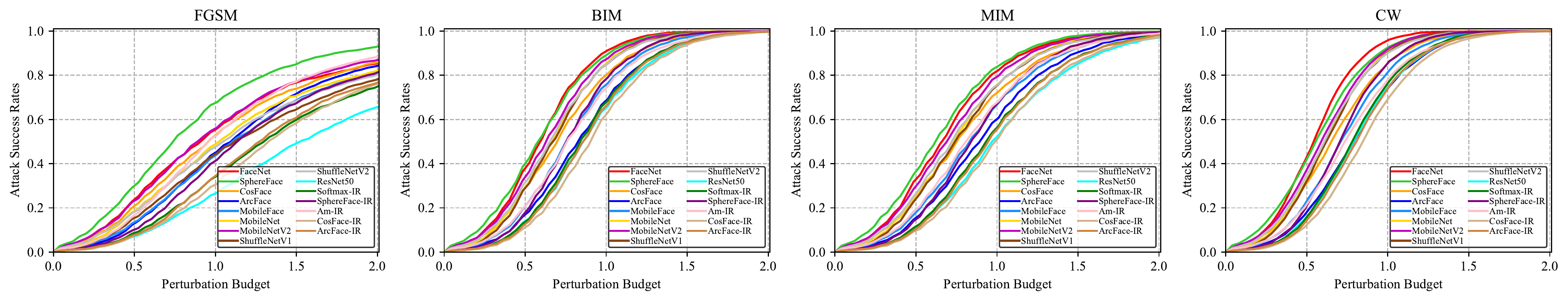}
\end{center}
\vspace{-2ex}
\caption{\emph{Asr vs. perturbation budget} curves of the $15$ models against dodging attacks under the $\ell_2$ norm.}
\label{fig:white-d-l2-asr-pert}
\vspace{-1ex}

\begin{center}
\includegraphics[width=0.99\linewidth]{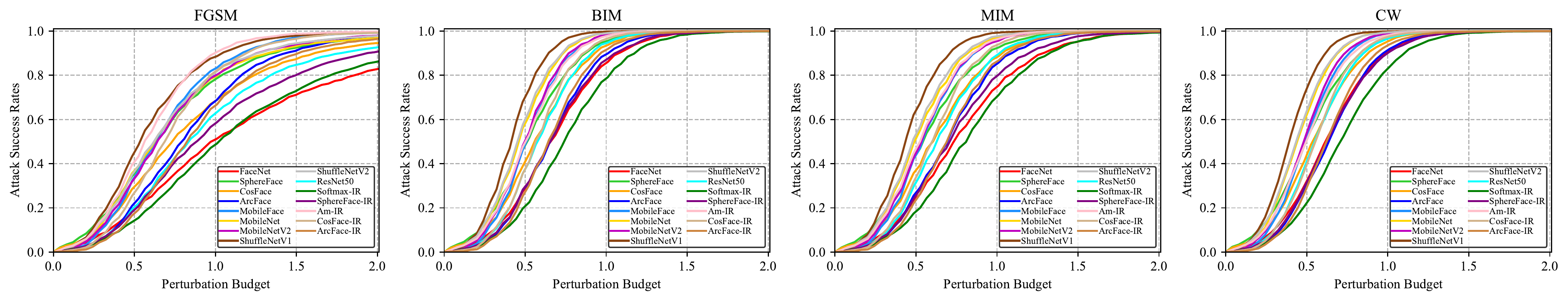}
\end{center}
\vspace{-3ex}
\caption{\textit{Asr vs. perturbation budget} curves of the $15$ models against impersonation attacks under the $\ell_2$ norm.}
\vspace{-3ex}
\label{fig:white-p-l2-asr-pert}
\end{figure*}


\vspace{-1ex}
\section{Evaluation Results}\label{sec:exp}
\vspace{-0.5ex}

Since the input size is different for each FR model, we adopt the normalized $\ell_2$ distance defined as $\bar{\ell}_2(\bm{a}) = {\|\bm{a}\|_2}/{\sqrt{d}}$ as the measurement for $\ell_2$ attacks, where $d$ is the dimension of a vector $\bm{a}$. 
We set $\alpha=1.5\epsilon/T$ for iterative methods, where $T$ is the maximum step. And we set $\mu=1.0$ for MIM and $c=10^{-3}$ for C\&W.  For LGC, the mask consists of four squares with a side length of $7$, and each square uses a randomly selected key point as the center. 
We compare LGC with a baseline called \emph{cutout iterative method} (CIM) that performs random cutout in Appendix {\color{red} C.2}, and perform ablation studies of LGC in Appendix {\color{red} C.3}. 
Due to the limited space, we only present the results on LFW. The results on commercial APIs and other datasets are listed in Appendix {\color{red} E} and Appendix {\color{red} F}, respectively.

\begin{table*}[t]
    \begin{center}
    \scriptsize
    \caption{
    The median distance of the \textit{minimum perturbations} of the $15$ models against dodging and impersonation attacks under the $\ell_2$ and $\ell_{\infty}$ norms.}
    \vspace{-1.5ex}
    \setlength{\tabcolsep}{5pt}
    \label{tab:median}
    \begin{tabular}{c||cc|cc|cc|cc||cc|cc|cc}
    \toprule
   \multirow{3}{*}{} & \multicolumn{8}{c||}{$\ell_2$} &
    \multicolumn{6}{c}{$\ell_\infty$}\\
    \cline{2-15}
     & \multicolumn{2}{c|}{FGSM} & \multicolumn{2}{c|}{BIM} & \multicolumn{2}{c|}{MIM} & \multicolumn{2}{c||}{C\&W} & \multicolumn{2}{c|}{FGSM} & \multicolumn{2}{c|}{BIM} & \multicolumn{2}{c}{MIM} \\
     \cline{2-15}
     & dod. & imp. & dod. & imp. & dod. & imp. & dod. & imp. & dod. & imp. & dod. & imp. & dod. & imp. \\
     \hline
     
    FaceNet & 0.92 & 1.17 & 0.59 & 0.66 & 0.66 & 0.73 & 0.54 & 0.61 & 1.75 & 2.14 & 1.17 & 1.31 & 1.28 & 1.42\\
    SphereFace & 0.73 & 0.64 & 0.58 & 0.52 & 0.62 & 0.55 & 0.56 & 0.50 & 1.31 & 1.11 & 1.06 & 0.92 & 1.12 & 0.97\\
    CosFace & 0.97 & 0.73 & 0.69 & 0.56 & 0.75 & 0.59 & 0.65 & 0.54 & 1.69 & 1.25 & 1.30 & 1.00 & 1.39 & 1.06\\
    ArcFace & 1.09 & 0.80 & 0.83 & 0.66 & 0.89 & 0.69 & 0.79 & 0.64 & 1.97 & 1.42 & 1.53 & 1.20 & 1.62 & 1.25\\\hline
    MobileFace & 1.11 & 0.62 & 0.75 & 0.50 & 0.83 & 0.53 & 0.71 & 0.49 & 1.98 & 1.12 & 1.44 & 0.94 & 1.55 & 0.97\\
    MobileNet & 1.03 & 0.64 & 0.67 & 0.47 & 0.73 & 0.50 & 0.62 & 0.44 & 1.75 & 1.08 & 1.22 & 0.83 & 1.31 & 0.88\\
    MobileNetV2 & 0.89 & 0.64 & 0.62 & 0.50 & 0.69 & 0.53 & 0.59 & 0.48 & 1.55 & 1.12 & 1.17 & 0.91 & 1.25 & 0.95\\
    ShuffleNetV1 & 1.12 & 0.53 & 0.67 & 0.41 & 0.75 & 0.44 & 0.62 & 0.39 & 2.02 & 0.97 & 1.28 & 0.77 & 1.39 & 0.80\\
    ShuffleNetV2 & 1.06 & 0.62 & 0.66 & 0.45 & 0.72 & 0.48 & 0.61 & 0.43 & 1.86 & 1.08 & 1.23 & 0.84 & 1.31 & 0.88\\
    ResNet50 & 1.53 & 0.84 & 0.86 & 0.58 & 0.97 & 0.62 & 0.79 & 0.55 & 2.64 & 1.44 & 1.59 & 1.05 & 1.75 & 1.11\\\hline
    Softmax-IR & 1.30 & 1.05 & 0.84 & 0.73 & 0.94 & 0.81 & 0.78 & 0.71 & 2.36 & 1.86 & 1.66 & 1.41 & 1.80 & 1.50\\
    SphereFace-IR & 1.14 & 0.90 & 0.77 & 0.66 & 0.84 & 0.70 & 0.70 & 0.63 & 2.12 & 1.62 & 1.50 & 1.25 & 1.62 & 1.31\\
    Am-IR & 0.95 & 0.56 & 0.77 & 0.50 & 0.81 & 0.52 & 0.74 & 0.49 & 1.67 & 1.00 & 1.41 & 0.92 & 1.47 & 0.94\\
    CosFace-IR & 1.33 & 0.69 & 0.91 & 0.56 & 0.98 & 0.59 & 0.85 & 0.55 & 2.45 & 1.27 & 1.78 & 1.09 & 1.91 & 1.14\\
    ArcFace-IR & 1.28 & 0.83 & 0.86 & 0.64 & 0.95 & 0.69 & 0.81 & 0.62 & 2.39 & 1.55 & 1.72 & 1.25 & 1.84 & 1.31\\
     \bottomrule
    \end{tabular}
    \end{center}
\vspace{-3ex}
\end{table*}


\begin{table*}[!htp]
    \begin{center}
    \scriptsize
    \setlength{\tabcolsep}{4.1pt}
    \caption{
    The median distance of the \textit{minimum perturbations} of different defense methods against dodging and impersonation attacks under the $\ell_2$ and $\ell_{\infty}$ norms.}
    \vspace{-1ex}
    \label{tab:median-defense}
    \begin{tabular}{c|c|cc|cc|cc|cc||cc|cc|cc}
    \toprule
   \multirow{3}{*}{} & \multirow{3}{*}{Clean} & \multicolumn{8}{c||}{$\ell_2$} &
    \multicolumn{6}{c}{$\ell_\infty$}\\
    \cline{3-16}
     && \multicolumn{2}{c|}{FGSM} & \multicolumn{2}{c|}{BIM} & \multicolumn{2}{c|}{MIM} & \multicolumn{2}{c||}{C\&W} & \multicolumn{2}{c|}{FGSM} & \multicolumn{2}{c|}{BIM} & \multicolumn{2}{c}{MIM} \\
     \cline{3-16}
     && dod. & imp. & dod. & imp. & dod. & imp. & dod. & imp. & dod. & imp. & dod. & imp. & dod. & imp. \\
     \hline
    JPEG~\cite{dziugaite2016study} & 99.6 & 2.46 & 2.00 & 1.36 & 1.16 & 1.31 & 1.12 & 0.75 & 0.67 & 4.19 & 3.31 & 2.77 & 2.23 & 2.58 & 2.16 \\
    Bit-Red~\cite{xu2018feature} & 99.6 & 1.37 & 1.06 & 0.86 & 0.73 & 0.92 & 0.78 & 0.78 & 0.74 & 3.00 & 2.00 & 2.00 & 1.12 & 2.00 & 1.67 \\
    R\&P~\cite{xie2017mitigating} & 99.4 & 4.50 & 8.01 & 2.05 & 2.20 & 2.28 & 2.50 & 1.86 & 2.02 & 8.00 & 10.31 & 4.56 & 4.77 & 4.25 & 4.59\\
    \hline
    PGD-AT~\cite{madry2018towards} &91.3& 4.14 & 3.95 & 2.37 & 2.37 & 2.70 & 2.70 & 2.06 & 2.14 & 12.94 & 12.38 & 10.34 & 10.23 & 10.83 & 10.62 \\
    TRADES~\cite{zhang2019theoretically} & 91.0 & 4.37 & 4.41 & 2.70 & 2.73 & 3.03 & 3.03 & 2.38 & 2.51 & 12.69 & 12.12 & 10.59 & 10.30 & 10.97 & 10.64\\
    \hline
	SphereFace-AT &88.9& 3.31 & 6.79 & 1.72 & 2.34 & 2.03 & 2.92 & 1.43 & 1.93 & 12.88 & 19.56 & 9.88 & 12.69 & 10.56 & 13.67 \\
	Am-AT & 85.0&3.78 & 4.25 & 2.31 & 2.64 & 2.61 & 2.92 & 2.00 & 2.41 & 11.62 & 12.50 & 9.88 & 10.64 & 10.20 & 11.02 \\
	CosFace-AT &86.2& 3.67 & 3.80 & 2.30 & 2.37 & 2.59 & 2.67 & 1.96 & 2.16 & 11.50 & 11.56 & 9.83 & 10.02 & 10.14 & 10.34 \\
	ArcFace-AT &87.7& 3.43 & 4.55 & 2.16 & 2.61 & 2.39 & 2.95 & 1.86 & 2.35 & 11.62 & 13.81 & 9.89 & 11.42 & 10.19 & 11.84

\\
     \bottomrule
    \end{tabular}
    \end{center}
    \vspace{-5ex}
\end{table*}
\vspace{-1ex}
\subsection{White-box Evaluation Results}
\vspace{-0.5ex}

We fixed the attack iterations as $20$ for BIM and MIM and $100$ for C\&W. To get minimum perturbation, a binary search with $10$ iterations is used after finding a feasible adversarial example by linear search. 

\textbf{Relationship between precision and robustness.} We show the results in Fig.~\ref{fig:white-d-l2-asr-pert}, Fig.~\ref{fig:white-p-l2-asr-pert}, and Tab.~\ref{tab:median}. Among attacks, C\&W attack gets the best performance, which accords with the performance on the general image classification~\cite{dong2020benchmarking,su2018robustness}.
However, we observe that those methods with the glorious precision, e.g., ArcFace and CosFace, do not have a positive effect on robustness, which is inconsistent with the conclusion of evaluating robustness for the ImageNet classification task where there exists a trade-off between precision and robustness.  Therefore, this is worth noting that learning more discriminative features from the metric space for precision cannot substantially improve robustness.

\textbf{Effects of different factors.} On the whole, model architecture is a more critical factor for robustness than loss functions. The models with a larger size of weights are more resistant to attacks under both norms. Thus selecting a proper larger network structure as a backbone usually leads to better resistance against adversarial examples, which also provides an insight for robustness.  

\textbf{Effects of different defenses.} 
For PGD-AT and TRADES, we set $\epsilon=8/255$, $\alpha=1/255$, and the number iteration as $9$ under $\ell_{\infty}$ norm on CASIA-WebFace~\cite{yi2014learning}. As shown in Tab.~\ref{tab:median-defense},  AT is the most robust method, which presents consistent performance on different adaptive loss constraints. TRADES performs slightly better than PGD-AT. Their robustness can generalize to the $\ell_{2}$ threat models when they are trained by the $\ell_{\infty}$. However, the overall performance against the $\ell_{2}$ threat models is inferior to the $\ell_{\infty}$ threat models. Besides, AT still results in a reduction of natural accuracy.

\begin{figure*}[t]
\centering
\includegraphics[width=0.9\linewidth]{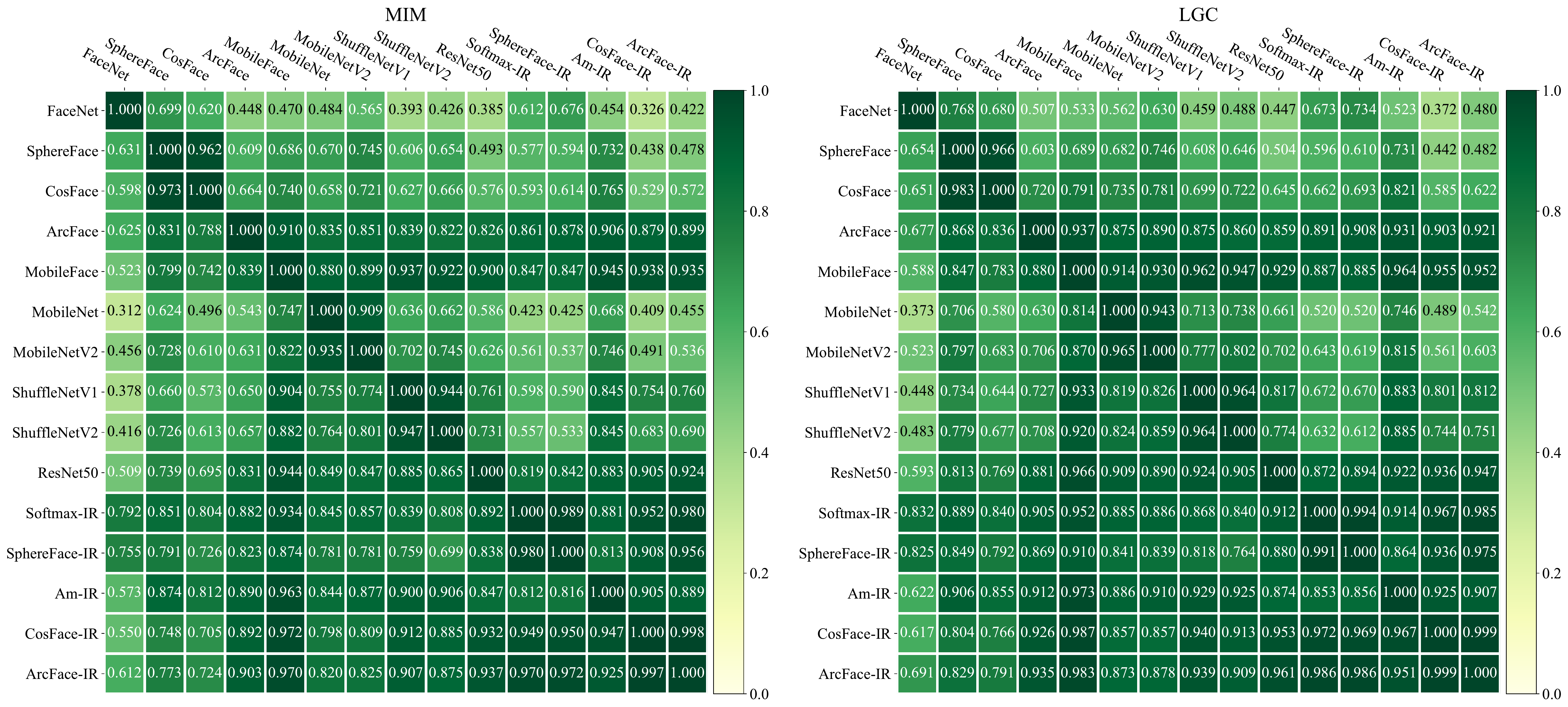}
\vspace{-2ex}
\caption{Asr of the $15$ models against black-box dodging attacks under the $\ell_{\infty}$ norm.}
\label{fig:heatmap_d}
\vspace{-2.5ex}
\end{figure*}
\begin{figure*}[t]
\centering
\includegraphics[width=0.9\linewidth]{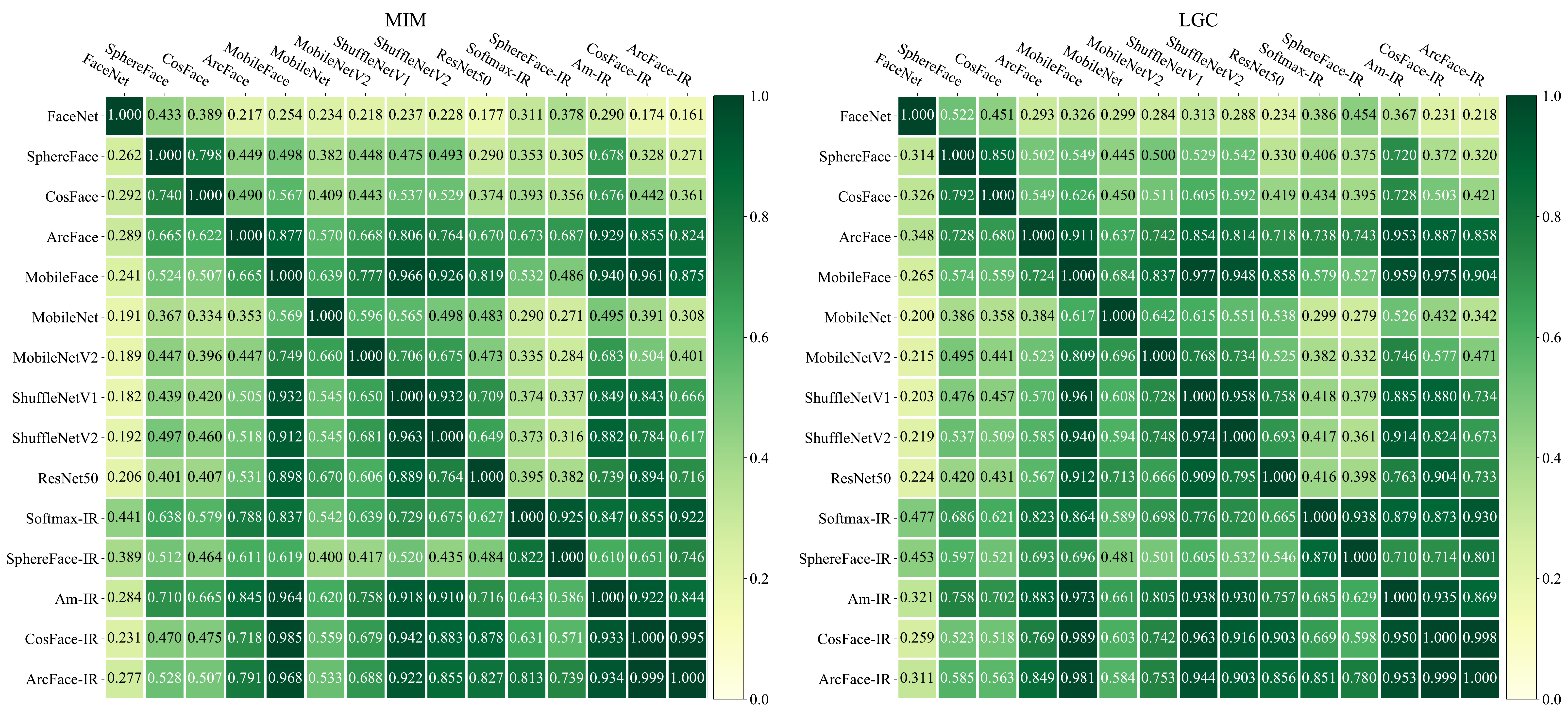}
\vspace{-2ex}
\caption{Asr of the $15$ models against black-box impersonation attacks under the $\ell_{\infty}$ norm.}
\label{fig:heatmap}
\vspace{-4ex}
\end{figure*}

\vspace{-1ex}
\subsection{Black-box Evaluation Results}
\vspace{-0.5ex}

For black-box attacks, we fix the perturbation budget as $\epsilon=8$ for $\ell_{\infty}$ attacks and $\epsilon=4$ for $\ell_2$ attacks. Considering that black-box attacks are harder than white-box attacks, we perform $100$ iterations for iterative attacks.
We choose a model as the substitute model to attack the others, and we show \textit{Asr} of the $15$ models against dodging and impersonation attacks under the $\ell_{\infty}$ norm in Fig.~\ref{fig:heatmap_d} and Fig.~\ref{fig:heatmap}.  


\textbf{Performance analysis.}
We observe that the same model has a similar overall transferable trend for different attack methods.  
Obviously, LGC improves the transferability of adversarial examples over FGSM, BIM, and MIM, resulting in higher Asr against the black-box models. Thus landmark-guided prior knowledge is beneficial for exploring transferable adversarial examples for the FR task.

\textbf{Robustness of different models.}
We observe that model architecture is a very important factor for robustness. Those models within the same architecture family have good transferability across each other. The models with light weights (e.g., MobileNet and ShuffleNet) are easily attacked by the adversarial examples generated against other models. One potential reason is that many networks (e.g., MobileNet and ShuffleNet) are based on improvements in the residual unit. 

\textbf{Robustness of different optimization loss.}
We find that the models among Am-IR, CosFace-IR and ArcFace have mutual transferability because their optimizations lie in angular space combined with additive margin.
FaceNet is the most difficult one to transfer to other models, which is trained on IncepetionResNetV1 based on the triplet loss in the Euclidean space.  This might provide new valuable insights for designing robust models against black-box attacks that selecting diverse backbones and train function will restrict the transferability of adversarial samples by common white-box FR models. 


\vspace{-1ex}
\subsection{Findings and Discussions}\label{sec:discussion}
\vspace{-0.5ex}

\textbf{First}, current mainstream models on FR, e.g., ArcFace and CosFace, have learned more discriminative features from the metric space, yet failing to substantially improve robustness. Targeted exploration of designing models is significantly taking into account robustness.
\textbf{Second}, we observe model architecture is a more critical factor. Light-weight models, e.g., MobileNet and ShuffleNet, have worse defensive performance than those of normal weights under all white-box and black-box settings. Thus selecting a suitable larger network structure as a backbone seems more effective for robustness.
\textbf{Third}, AT is the most robust method among the defense methods studied in this paper,  which presents consistent performance on different adaptive face loss constraints. However, the robustness under the $\ell_{2}$ norms is inferior to the $\ell_{\infty}$ norms that are used in the training phase. Besides, AT still results in a reduction of natural accuracy and high training cost in FR.
\textbf{Fourth},  we observe that the models with similar structures have higher transferability. Besides, similar training loss functions are also mutually transferable, such as LCML, ArgFace, and Am-Softmax.
\textbf{Fifth}, the attack performance under $\ell_2$ and $\ell_\infty$ norms is very similar in generalization ability for those models belonging to the normal training. Besides, the difficulty of impersonation attacks is higher than that of dodging attacks.
\textbf{Sixth}, the proposed LGC makes the generated adversarial examples more transferable based on the characteristics of face tasks, which can be flexibly applied to many black-box attack methods, such as MIM, DIM and TIM.
\textbf{Seventh}, similar overall robustness conclusions across different testing datasets are presented for the same FR models, including the widely used LFW, YTF, CFP-FP and MegaFace.

\vspace{-1ex}
\section{Conclusion}
\vspace{-0.5ex}

To the best of our knowledge, we performed the largest experiment on evaluating the adversarial robustness on FR, covering diverse FR models against both white-box and black-box attacks. We conjecture that robustness should also be considered in model designs while pursuing higher accuracy. We also provide thorough evaluation metrics and valuable insights on robust FR of future works, and even in other metric learning tasks such as image retrieval, objective re-identification, and so on. 

\clearpage
{\small
\bibliographystyle{plainnat}
\bibliography{egbib}}

\begin{thebibliography}{61}
\providecommand{\natexlab}[1]{#1}
\providecommand{\url}[1]{\texttt{#1}}
\expandafter\ifx\csname urlstyle\endcsname\relax
  \providecommand{\doi}[1]{doi: #1}\else
  \providecommand{\doi}{doi: \begingroup \urlstyle{rm}\Url}\fi

\bibitem[Brendel et~al.(2020)Brendel, Rauber, Kurakin, Papernot, Veliqi,
  Mohanty, Laurent, Salath{\'e}, Bethge, Yu, et~al.]{brendel2020adversarial}
Wieland Brendel, Jonas Rauber, Alexey Kurakin, Nicolas Papernot, Behar Veliqi,
  Sharada~P Mohanty, Florian Laurent, Marcel Salath{\'e}, Matthias Bethge,
  Yaodong Yu, et~al.
\newblock Adversarial vision challenge.
\newblock In \emph{The NeurIPS'18 Competition}, pages 129--153. Springer, 2020.

\bibitem[Carlini and Wagner(2017)]{carlini2017towards}
Nicholas Carlini and David Wagner.
\newblock Towards evaluating the robustness of neural networks.
\newblock In \emph{IEEE Symposium on Security and Privacy}, 2017.

\bibitem[Carlini et~al.(2019)Carlini, Athalye, Papernot, Brendel, Rauber,
  Tsipras, Goodfellow, and Madry]{carlini2019evaluating}
Nicholas Carlini, Anish Athalye, Nicolas Papernot, Wieland Brendel, Jonas
  Rauber, Dimitris Tsipras, Ian Goodfellow, and Aleksander Madry.
\newblock On evaluating adversarial robustness.
\newblock \emph{arXiv preprint arXiv:1902.06705}, 2019.

\bibitem[Chen et~al.(2018)Chen, Liu, Gao, and Han]{chen2018mobilefacenets}
Sheng Chen, Yang Liu, Xiang Gao, and Zhen Han.
\newblock Mobilefacenets: Efficient cnns for accurate real-time face
  verification on mobile devices.
\newblock In \emph{Chinese Conference on Biometric Recognition}, pages
  428--438. Springer, 2018.

\bibitem[Deng et~al.(2019{\natexlab{a}})Deng, Guo, Xue, and
  Zafeiriou]{deng2018arcface}
Jiankang Deng, Jia Guo, Niannan Xue, and Stefanos Zafeiriou.
\newblock Arcface: Additive angular margin loss for deep face recognition.
\newblock In \emph{The IEEE Conference on Computer Vision and Pattern
  Recognition (CVPR)}, 2019{\natexlab{a}}.

\bibitem[Deng et~al.(2019{\natexlab{b}})Deng, Guo, Xue, and
  Zafeiriou]{deng2019arcface}
Jiankang Deng, Jia Guo, Niannan Xue, and Stefanos Zafeiriou.
\newblock Arcface: Additive angular margin loss for deep face recognition.
\newblock In \emph{Proceedings of the IEEE Conference on Computer Vision and
  Pattern Recognition}, pages 4690--4699, 2019{\natexlab{b}}.

\bibitem[DeVries and Taylor(2017)]{devries2017improved}
Terrance DeVries and Graham~W Taylor.
\newblock Improved regularization of convolutional neural networks with cutout.
\newblock \emph{arXiv preprint arXiv:1708.04552}, 2017.

\bibitem[Dong et~al.(2018)Dong, Liao, Pang, Su, Zhu, Hu, and Li]{Dong2017}
Yinpeng Dong, Fangzhou Liao, Tianyu Pang, Hang Su, Jun Zhu, Xiaolin Hu, and
  Jianguo Li.
\newblock Boosting adversarial attacks with momentum.
\newblock In \emph{Proceedings of the IEEE Conference on Computer Vision and
  Pattern Recognition (CVPR)}, 2018.

\bibitem[Dong et~al.(2019{\natexlab{a}})Dong, Pang, Su, and
  Zhu]{dong2019evading}
Yinpeng Dong, Tianyu Pang, Hang Su, and Jun Zhu.
\newblock Evading defenses to transferable adversarial examples by
  translation-invariant attacks.
\newblock In \emph{Proceedings of the IEEE Conference on Computer Vision and
  Pattern Recognition (CVPR)}, 2019{\natexlab{a}}.

\bibitem[Dong et~al.(2019{\natexlab{b}})Dong, Su, Wu, Li, Liu, Zhang, and
  Zhu]{dong2019efficient}
Yinpeng Dong, Hang Su, Baoyuan Wu, Zhifeng Li, Wei Liu, Tong Zhang, and Jun
  Zhu.
\newblock Efficient decision-based black-box adversarial attacks on face
  recognition.
\newblock In \emph{Proceedings of the IEEE Conference on Computer Vision and
  Pattern Recognition (CVPR)}, 2019{\natexlab{b}}.

\bibitem[Dong et~al.(2020)Dong, Fu, Yang, Pang, Su, Xiao, and
  Zhu]{dong2020benchmarking}
Yinpeng Dong, Qi-An Fu, Xiao Yang, Tianyu Pang, Hang Su, Zihao Xiao, and Jun
  Zhu.
\newblock Benchmarking adversarial robustness on image classification.
\newblock In \emph{Proceedings of the IEEE/CVF Conference on Computer Vision
  and Pattern Recognition}, pages 321--331, 2020.

\bibitem[Dziugaite et~al.(2016)Dziugaite, Ghahramani, and
  Roy]{dziugaite2016study}
Gintare~Karolina Dziugaite, Zoubin Ghahramani, and Daniel~M Roy.
\newblock A study of the effect of jpg compression on adversarial images.
\newblock \emph{arXiv preprint arXiv:1608.00853}, 2016.

\bibitem[Goodfellow et~al.(2015)Goodfellow, Shlens, and
  Szegedy]{goodfellow2014explaining}
Ian~J Goodfellow, Jonathon Shlens, and Christian Szegedy.
\newblock Explaining and harnessing adversarial examples.
\newblock In \emph{International Conference on Learning Representations
  (ICLR)}, 2015.

\bibitem[Gutta et~al.(2002)Gutta, Philomin, and
  Trajkovic]{gutta2002investigation}
Srinivas Gutta, Vasanth Philomin, and Miroslav Trajkovic.
\newblock An investigation into the use of partial-faces for face recognition.
\newblock In \emph{fgr}, pages 33--38, 2002.

\bibitem[He et~al.(2016{\natexlab{a}})He, Zhang, Ren, and Sun]{he2015deep}
Kaiming He, Xiangyu Zhang, Shaoqing Ren, and Jian Sun.
\newblock Deep residual learning for image recognition.
\newblock In \emph{CVPR}, 2016{\natexlab{a}}.

\bibitem[He et~al.(2016{\natexlab{b}})He, Zhang, Ren, and Sun]{he2016}
Kaiming He, Xiangyu Zhang, Shaoqing Ren, and Jian Sun.
\newblock Identity mappings in deep residual networks.
\newblock In \emph{European Conference on Computer Vision (ECCV)}, pages
  630--645. Springer, 2016{\natexlab{b}}.

\bibitem[Howard et~al.(2017)Howard, Zhu, Chen, Kalenichenko, Wang, Weyand,
  Andreetto, and Adam]{howard2017mobilenets}
Andrew~G Howard, Menglong Zhu, Bo~Chen, Dmitry Kalenichenko, Weijun Wang,
  Tobias Weyand, Marco Andreetto, and Hartwig Adam.
\newblock Mobilenets: Efficient convolutional neural networks for mobile vision
  applications.
\newblock \emph{arXiv preprint arXiv:1704.04861}, 2017.

\bibitem[Hu et~al.(2015)Hu, Yang, Yi, Kittler, Christmas, Li, and
  Hospedales]{hu2015face}
Guosheng Hu, Yongxin Yang, Dong Yi, Josef Kittler, William Christmas, Stan~Z
  Li, and Timothy Hospedales.
\newblock When face recognition meets with deep learning: an evaluation of
  convolutional neural networks for face recognition.
\newblock In \emph{Proceedings of the IEEE international conference on computer
  vision workshops}, pages 142--150, 2015.

\bibitem[Huang et~al.(2007)Huang, Mattar, Berg, and
  Learned-Miller]{huang2008labeled}
Gary~B Huang, Marwan Mattar, Tamara Berg, and Eric Learned-Miller.
\newblock Labeled faces in the wild: A database forstudying face recognition in
  unconstrained environments.
\newblock In \emph{Technical report}, 2007.

\bibitem[Kemelmacher-Shlizerman et~al.(2016)Kemelmacher-Shlizerman, Seitz,
  Miller, and Brossard]{kemelmacher2016megaface}
Ira Kemelmacher-Shlizerman, Steven~M Seitz, Daniel Miller, and Evan Brossard.
\newblock The megaface benchmark: 1 million faces for recognition at scale.
\newblock In \emph{Proceedings of the IEEE conference on computer vision and
  pattern recognition}, pages 4873--4882, 2016.

\bibitem[Kingma and Ba(2015)]{Kingma2014}
Diederik Kingma and Jimmy Ba.
\newblock Adam: A method for stochastic optimization.
\newblock In \emph{International Conference on Learning Representations
  (ICLR)}, 2015.

\bibitem[Kurakin et~al.(2017{\natexlab{a}})Kurakin, Goodfellow, and
  Bengio]{Kurakin2016}
Alexey Kurakin, Ian Goodfellow, and Samy Bengio.
\newblock Adversarial examples in the physical world.
\newblock In \emph{International Conference on Learning Representations (ICLR)
  Workshops}, 2017{\natexlab{a}}.

\bibitem[Kurakin et~al.(2017{\natexlab{b}})Kurakin, Goodfellow, and
  Bengio]{kurakin2016adversarial}
Alexey Kurakin, Ian Goodfellow, and Samy Bengio.
\newblock Adversarial machine learning at scale.
\newblock In \emph{International Conference on Learning Representations
  (ICLR)}, 2017{\natexlab{b}}.

\bibitem[Kurakin et~al.(2018)Kurakin, Goodfellow, Bengio, Dong, Liao, Liang,
  Pang, Zhu, Hu, Xie, et~al.]{kurakin2018competation}
Alexey Kurakin, Ian Goodfellow, Samy Bengio, Yinpeng Dong, Fangzhou Liao, Ming
  Liang, Tianyu Pang, Jun Zhu, Xiaolin Hu, Cihang Xie, et~al.
\newblock Adversarial attacks and defences competition.
\newblock \emph{arXiv preprint arXiv:1804.00097}, 2018.

\bibitem[Liu et~al.(2017)Liu, Wen, Yu, Li, Raj, and Song]{liu2017sphereface}
Weiyang Liu, Yandong Wen, Zhiding Yu, Ming Li, Bhiksha Raj, and Le~Song.
\newblock Sphereface: Deep hypersphere embedding for face recognition.
\newblock In \emph{CVPR}, 2017.

\bibitem[Ma et~al.(2018)Ma, Zhang, Zheng, and Sun]{ma2018shufflenet}
Ningning Ma, Xiangyu Zhang, Hai-Tao Zheng, and Jian Sun.
\newblock Shufflenet v2: Practical guidelines for efficient cnn architecture
  design.
\newblock In \emph{Proceedings of the European Conference on Computer Vision
  (ECCV)}, pages 116--131, 2018.

\bibitem[Madry et~al.(2018{\natexlab{a}})Madry, Makelov, Schmidt, Tsipras, and
  Vladu]{madry2017towards}
Aleksander Madry, Aleksandar Makelov, Ludwig Schmidt, Dimitris Tsipras, and
  Adrian Vladu.
\newblock Towards deep learning models resistant to adversarial attacks.
\newblock In \emph{International Conference on Learning Representations
  (ICLR)}, 2018{\natexlab{a}}.

\bibitem[Madry et~al.(2018{\natexlab{b}})Madry, Makelov, Schmidt, Tsipras, and
  Vladu]{madry2018towards}
Aleksander Madry, Aleksandar Makelov, Ludwig Schmidt, Dimitris Tsipras, and
  Adrian Vladu.
\newblock Towards deep learning models resistant to adversarial attacks.
\newblock In \emph{International Conference on Learning Representations
  (ICLR)}, 2018{\natexlab{b}}.

\bibitem[Papernot et~al.(2016{\natexlab{a}})Papernot, Faghri, Carlini,
  Goodfellow, Feinman, Kurakin, Xie, Sharma, Brown, Roy,
  et~al.]{papernot2016technical}
Nicolas Papernot, Fartash Faghri, Nicholas Carlini, Ian Goodfellow, Reuben
  Feinman, Alexey Kurakin, Cihang Xie, Yash Sharma, Tom Brown, Aurko Roy,
  et~al.
\newblock Technical report on the cleverhans v2. 1.0 adversarial examples
  library.
\newblock \emph{arXiv preprint arXiv:1610.00768}, 2016{\natexlab{a}}.

\bibitem[Papernot et~al.(2016{\natexlab{b}})Papernot, McDaniel, Goodfellow,
  Jha, Celik, and Swami]{Papernot2016}
Nicolas Papernot, Patrick McDaniel, Ian Goodfellow, Somesh Jha, Z~Berkay Celik,
  and Ananthram Swami.
\newblock Practical black-box attacks against deep learning systems using
  adversarial examples.
\newblock \emph{arXiv preprint arXiv:1602.02697}, 2016{\natexlab{b}}.

\bibitem[Rauber et~al.(2017)Rauber, Brendel, and Bethge]{rauber2017foolbox}
Jonas Rauber, Wieland Brendel, and Matthias Bethge.
\newblock Foolbox v0. 8.0: A python toolbox to benchmark the robustness of
  machine learning models.
\newblock \emph{arXiv preprint arXiv:1707.04131}, 2017.

\bibitem[Sandler et~al.(2018)Sandler, Howard, Zhu, Zhmoginov, and
  Chen]{sandler2018mobilenetv2}
Mark Sandler, Andrew Howard, Menglong Zhu, Andrey Zhmoginov, and Liang-Chieh
  Chen.
\newblock Mobilenetv2: Inverted residuals and linear bottlenecks.
\newblock In \emph{Proceedings of the IEEE Conference on Computer Vision and
  Pattern Recognition}, pages 4510--4520, 2018.

\bibitem[Schroff et~al.(2015)Schroff, Kalenichenko, and
  Philbin]{schroff2015facenet}
Florian Schroff, Dmitry Kalenichenko, and James Philbin.
\newblock Facenet: A unified embedding for face recognition and clustering.
\newblock In \emph{CVPR}, 2015.

\bibitem[Selvaraju et~al.(2017)Selvaraju, Cogswell, Das, Vedantam, Parikh, and
  Batra]{selvaraju2017grad}
Ramprasaath~R Selvaraju, Michael Cogswell, Abhishek Das, Ramakrishna Vedantam,
  Devi Parikh, and Dhruv Batra.
\newblock Grad-cam: Visual explanations from deep networks via gradient-based
  localization.
\newblock In \emph{Proceedings of the IEEE international conference on computer
  vision}, pages 618--626, 2017.

\bibitem[Sengupta et~al.(2016)Sengupta, Chen, Castillo, Patel, Chellappa, and
  Jacobs]{Sengupta2016Frontal}
Soumyadip Sengupta, Jun~Cheng Chen, Carlos Castillo, Vishal~M. Patel, Rama
  Chellappa, and David~W. Jacobs.
\newblock Frontal to profile face verification in the wild.
\newblock In \emph{2016 IEEE Winter Conference on Applications of Computer
  Vision (WACV)}, 2016.

\bibitem[Sharif et~al.(2016)Sharif, Bhagavatula, Bauer, and
  Reiter]{sharif2016accessorize}
Mahmood Sharif, Sruti Bhagavatula, Lujo Bauer, and Michael~K Reiter.
\newblock Accessorize to a crime: Real and stealthy attacks on state-of-the-art
  face recognition.
\newblock In \emph{Proceedings of the 2016 ACM SIGSAC Conference on Computer
  and Communications Security}, pages 1528--1540. ACM, 2016.

\bibitem[Sharif et~al.(2017)Sharif, Bhagavatula, and
  Bauer]{sharif2017adversarial}
Mahmood Sharif, Sruti Bhagavatula, and Bauer.
\newblock Adversarial generative nets: Neural network attacks on
  state-of-the-art face recognition.
\newblock \emph{arXiv preprint arXiv:1801.00349}, 2017.

\bibitem[Simonyan and Zisserman(2015)]{simonyan2014very}
Karen Simonyan and Andrew Zisserman.
\newblock Very deep convolutional networks for large-scale image recognition.
\newblock In \emph{International Conference on Learning Representations
  (ICLR)}, 2015.

\bibitem[Su et~al.(2018)Su, Zhang, Chen, Yi, Chen, and Gao]{su2018robustness}
Dong Su, Huan Zhang, Hongge Chen, Jinfeng Yi, Pin-Yu Chen, and Yupeng Gao.
\newblock Is robustness the cost of accuracy?--a comprehensive study on the
  robustness of 18 deep image classification models.
\newblock In \emph{Proceedings of the European Conference on Computer Vision
  (ECCV)}, pages 631--648, 2018.

\bibitem[Sun et~al.(2014)Sun, Chen, Wang, and Tang]{sun2014deep}
Yi~Sun, Yuheng Chen, Xiaogang Wang, and Xiaoou Tang.
\newblock Deep learning face representation by joint
  identification-verification.
\newblock In \emph{Advances in neural information processing systems}, pages
  1988--1996, 2014.

\bibitem[Szegedy et~al.(2014)Szegedy, Zaremba, Sutskever, Bruna, Erhan,
  Goodfellow, and Fergus]{szegedy2013intriguing}
Christian Szegedy, Wojciech Zaremba, Ilya Sutskever, Joan Bruna, Dumitru Erhan,
  Ian Goodfellow, and Rob Fergus.
\newblock Intriguing properties of neural networks.
\newblock In \emph{International Conference on Learning Representations
  (ICLR)}, 2014.

\bibitem[Szegedy et~al.(2015)Szegedy, Liu, Jia, Sermanet, Reed, Anguelov,
  Erhan, Vanhoucke, and Rabinovich]{szegedy2015going}
Christian Szegedy, Wei Liu, Yangqing Jia, Pierre Sermanet, Scott Reed, Dragomir
  Anguelov, Dumitru Erhan, Vincent Vanhoucke, and Andrew Rabinovich.
\newblock Going deeper with convolutions.
\newblock In \emph{CVPR}, 2015.

\bibitem[Taigman et~al.(2014)Taigman, Yang, Ranzato, and
  Wolf]{taigman2014deepface}
Yaniv Taigman, Ming Yang, Marc'Aurelio Ranzato, and Lior Wolf.
\newblock Deepface: Closing the gap to human-level performance in face
  verification.
\newblock In \emph{CVPR}, 2014.

\bibitem[Tong et~al.(2021)Tong, Chen, Ni, Cheng, Song, Chen, and
  Vorobeychik]{tong2021facesec}
Liang Tong, Zhengzhang Chen, Jingchao Ni, Wei Cheng, Dongjin Song, Haifeng
  Chen, and Yevgeniy Vorobeychik.
\newblock Facesec: A fine-grained robustness evaluation framework for face
  recognition systems.
\newblock In \emph{Proceedings of the IEEE/CVF Conference on Computer Vision
  and Pattern Recognition}, pages 13254--13263, 2021.

\bibitem[Tsipras et~al.(2018)Tsipras, Santurkar, Engstrom, Turner, and
  Madry]{tsipras2018robustness}
Dimitris Tsipras, Shibani Santurkar, Logan Engstrom, Alexander Turner, and
  Aleksander Madry.
\newblock Robustness may be at odds with accuracy.
\newblock \emph{arXiv preprint arXiv:1805.12152}, 2018.

\bibitem[Wang et~al.(2018{\natexlab{a}})Wang, Cheng, Liu, and
  Liu]{wang2018additive}
Feng Wang, Jian Cheng, Weiyang Liu, and Haijun Liu.
\newblock Additive margin softmax for face verification.
\newblock \emph{IEEE Signal Processing Letters}, 25\penalty0 (7):\penalty0
  926--930, 2018{\natexlab{a}}.

\bibitem[Wang et~al.(2018{\natexlab{b}})Wang, Wang, Zhou, Ji, Li, Gong, Zhou,
  and Liu]{wang2018cosface}
Hao Wang, Yitong Wang, Zheng Zhou, Xing Ji, Zhifeng Li, Dihong Gong, Jingchao
  Zhou, and Wei Liu.
\newblock Cosface: Large margin cosine loss for deep face recognition.
\newblock In \emph{CVPR}, 2018{\natexlab{b}}.

\bibitem[Wen et~al.(2016)Wen, Zhang, Li, and Qiao]{wen2016discriminative}
Yandong Wen, Kaipeng Zhang, Zhifeng Li, and Yu~Qiao.
\newblock A discriminative feature learning approach for deep face recognition.
\newblock In \emph{ECCV}, 2016.

\bibitem[Wolf et~al.(2011)Wolf, Hassner, and Maoz]{wolf2011face}
Lior Wolf, Tal Hassner, and Itay Maoz.
\newblock Face recognition in unconstrained videos with matched background
  similarity.
\newblock In \emph{The IEEE Conference on Computer Vision and Pattern
  Recognition (CVPR)}, 2011.

\bibitem[Xie et~al.(2018)Xie, Wang, Zhang, Ren, and Yuille]{xie2017mitigating}
Cihang Xie, Jianyu Wang, Zhishuai Zhang, Zhou Ren, and Alan Yuille.
\newblock Mitigating adversarial effects through randomization.
\newblock In \emph{International Conference on Learning Representations
  (ICLR)}, 2018.

\bibitem[Xie et~al.(2019)Xie, Zhang, Zhou, Bai, Wang, Ren, and
  Yuille]{xie2019improving}
Cihang Xie, Zhishuai Zhang, Yuyin Zhou, Song Bai, Jianyu Wang, Zhou Ren, and
  Alan~L Yuille.
\newblock Improving transferability of adversarial examples with input
  diversity.
\newblock In \emph{Proceedings of the IEEE Conference on Computer Vision and
  Pattern Recognition (CVPR)}, 2019.

\bibitem[Xu et~al.(2018)Xu, Evans, and Qi]{xu2018feature}
Weilin Xu, David Evans, and Yanjun Qi.
\newblock Feature squeezing: Detecting adversarial examples in deep neural
  networks.
\newblock In \emph{Proceedings of the Network and Distributed System Security
  Symposium (NDSS)}, 2018.

\bibitem[Yang et~al.(2019)Yang, Luo, Bao, Gao, Gong, Zheng, Li, and
  Liu]{yang2019face}
Xiao Yang, Wenhan Luo, Linchao Bao, Yuan Gao, Dihong Gong, Shibao Zheng,
  Zhifeng Li, and Wei Liu.
\newblock Face anti-spoofing: Model matters, so does data.
\newblock In \emph{Proceedings of the IEEE Conference on Computer Vision and
  Pattern Recognition}, pages 3507--3516, 2019.

\bibitem[Yang et~al.(2020)Yang, Wei, Zhang, and Zhu]{yang2020design}
Xiao Yang, Fangyun Wei, Hongyang Zhang, and Jun Zhu.
\newblock Design and interpretation of universal adversarial patches in face
  detection.
\newblock In \emph{Computer Vision--ECCV 2020: 16th European Conference,
  Glasgow, UK, August 23--28, 2020, Proceedings, Part XVII 16}, pages 174--191.
  Springer, 2020.

\bibitem[Yang et~al.(2021)Yang, Dong, Pang, Su, and Zhu]{yang2021boosting}
Xiao Yang, Yinpeng Dong, Tianyu Pang, Hang Su, and Jun Zhu.
\newblock Boosting transferability of targeted adversarial examples via
  hierarchical generative networks.
\newblock \emph{arXiv preprint arXiv:2107.01809}, 2021.

\bibitem[Yi et~al.(2014)Yi, Lei, Liao, and Li]{yi2014learning}
Dong Yi, Zhen Lei, Shengcai Liao, and Stan~Z Li.
\newblock Learning face representation from scratch.
\newblock \emph{arXiv preprint arXiv:1411.7923}, 2014.

\bibitem[Zhang et~al.(2019)Zhang, Yu, Jiao, Xing, Ghaoui, and
  Jordan]{zhang2019theoretically}
Hongyang Zhang, Yaodong Yu, Jiantao Jiao, Eric~P Xing, Laurent~El Ghaoui, and
  Michael~I Jordan.
\newblock Theoretically principled trade-off between robustness and accuracy.
\newblock In \emph{International Conference on Machine Learning (ICML)}, 2019.

\bibitem[Zhang et~al.(2016)Zhang, Zhang, Li, and Qiao]{zhang2016joint}
Kaipeng Zhang, Zhanpeng Zhang, Zhifeng Li, and Yu~Qiao.
\newblock Joint face detection and alignment using multitask cascaded
  convolutional networks.
\newblock \emph{IEEE Signal Processing Letters}, 23\penalty0 (10):\penalty0
  1499--1503, 2016.

\bibitem[Zhang et~al.(2018)Zhang, Zhou, Lin, and Sun]{zhang2018shufflenet}
Xiangyu Zhang, Xinyu Zhou, Mengxiao Lin, and Jian Sun.
\newblock Shufflenet: An extremely efficient convolutional neural network for
  mobile devices.
\newblock In \emph{Proceedings of the IEEE Conference on Computer Vision and
  Pattern Recognition}, pages 6848--6856, 2018.

\bibitem[Zheng and Deng(2018)]{zheng2018cross}
Tianyue Zheng and Weihong Deng.
\newblock Cross-pose lfw: A database for studying cross-pose face recognition
  in unconstrained environments.
\newblock \emph{Beijing University of Posts and Telecommunications, Tech. Rep},
  5, 2018.

\bibitem[Zheng et~al.(2017)Zheng, Deng, and Hu]{zheng2017cross}
Tianyue Zheng, Weihong Deng, and Jiani Hu.
\newblock Cross-age lfw: A database for studying cross-age face recognition in
  unconstrained environments.
\newblock \emph{arXiv preprint arXiv:1708.08197}, 2017.

\end{thebibliography}

\appendix

\section{Related Adversarial Robustness Platforms}

The mainstream libraries for adversarial machine learning include CleverHans~\cite{papernot2016technical}, Foolbox~\cite{rauber2017foolbox} and RealSafe~\cite{dong2020benchmarking}, which focus on image classification. These
libraries or platforms cannot support our evaluations in FR in terms of attacks, defenses and evaluations. Therefore, we define the modified attack formulations and representative defenses in FR, to implement the robustness evaluation. Another related work to ours is FACESEC~\cite{tong2021facesec}), Specifically, FACESEC focuses on $\ell_{0}$-norm attacks involving sticker attacks against 5 naturally FR models. As a comparison, our work has differences in two main aspects: 1) we aim to perform a comprehensive robustness evaluation in FR to facilitate a better understanding of the adversarial vulnerability. Thus we try our best to involve more diverse FR models, including 15 popular \emph{naturally trained} FR models, 9 models with representative \emph{defense mechanisms} and 2 commercial API services; 2) FACESEC adopts adversarial patch attacks under $\ell_{0}$-norm constraint, the demos of which in the physical world are very convenient. However, adversarial patch attacks are easily affected by the position, shape and scale of the adversarial patch for different individual data points, thereby hard to guarantee the fairness and effectiveness when making a general robustness evaluation based on \emph{large-scale} facial images with various pose, expression, and illuminations. We mainly consider more popular $\ell_{2}$ and $\ell_{\infty}$ threat models than $\ell_{0}$-norm ones, which are appropriate and widely applied in the robustness evaluation works~\cite{papernot2016technical,rauber2017foolbox,dong2020benchmarking}.  To the best of our knowledge, we perform the \emph{largest} experiment on evaluating the adversarial robustness on face recognition, and present many key findings and valuable insights in multiple dimensions.

\section{Background}
In this section, we overview the research efforts on loss functions, network architectures and different datasets in FR.
\subsection{Loss Function on FR} 
\textbf{Euclidean-distance-based loss.} Softmax loss is commonly adopted in image classification which encourages the separability of features. However, FR requires more discriminative features that intra-variations are larger than inter-differences. Owing to the development of deep CNNs, FR has obtained remarkable progress recently.
DeepFace~\cite{taigman2014deepface} and DeepID~\cite{sun2014deep} regard FR as a multi-class classification task and train deep CNNs supervised by the softmax loss.
The evolved methods~\cite{sun2014deep,schroff2015facenet,wen2016discriminative} treat FR as a metric learning problem and learn highly discriminative features by Euclidean-distance-based loss.
The contrastive loss~\cite{sun2014deep} and triplet loss~\cite{schroff2015facenet} are proposed to increase the Euclidean margin in the feature space between different classes. Center loss~\cite{wen2016discriminative} aims to reduce intra-variance by assigning a learned center for each class.

\textbf{Angular-margin-based loss.} With the advanced training techniques, angular-margin-based loss~\cite{liu2017sphereface,wang2018cosface,deng2018arcface} is proposed to achieve the separability among features with a larger angular distance. SphereFace~\cite{liu2017sphereface} first introduces the angular softmax loss (A-Softmax). To overcome the optimization difficulty of A-Softmax that introduces the angular margin in a multiplicative manner, CosFace~\cite{wang2018cosface}, AM-Softmax~
\cite{wang2018additive} and ArcFace~\cite{deng2018arcface} adopt the additive angular margin loss to improve the margin in the angular space for achieving good performance, which are easy to implement and converge without many tricky hyperparameters.

\subsection{Network Architecture on FR}
\textbf{Mainstream architectures.} As advanced network architectures of image classification are constantly improved, the networks in FR are also developing gradually. The typical CNN architectures, including VGGNet~\cite{simonyan2014very}, GoogleNet~\cite{szegedy2015going} and ResNet~\cite{he2015deep}, 
are widely introduced as the baselines in FR after directly employed or slightly modified. DeepFace~\cite{taigman2014deepface} adopts CNNs with several locally connected layers, and FaceNet~\cite{schroff2015facenet} achieves good performance on LFW by using a triplet loss function based on GoogleNet architecture. SphereFace~\cite{liu2017sphereface} proposed a modified ResNet architecture to boost the adaptiveness in FR.

\textbf{Light-weight networks.} Deep CNNs with plenty of layers and millions of parameters can achieve good accuracy, yet requiring large computing resource and training cost. To tackle this problem, MobileFace~\cite{chen2018mobilefacenets} introduces downsampling and bottleneck residual block, achieving acceptable performance on the LFW dataset. Besides, some other light-weight CNNs in image classification, including MobileNet~\cite{howard2017mobilenets,sandler2018mobilenetv2} and ShuffleNet~\cite{zhang2018shufflenet,ma2018shufflenet}, pay more attention on the adaptiveness and effectiveness in FR.

\subsection{Datasets} The Labeled Face in the Wild (\textbf{LFW}) dataset is the most widely used benchmark for face verification on images, which contains $13,233$ face images from $5,749$ different individuals. LFW includes faces with various pose, expression, and illuminations. The \emph{unrestricted with labeled outside data} protocol includes $6,000$ face pairs, including $3,000$ pairs with the same identities and $3,000$ pairs with different identities. YouTube Faces Database (\textbf{YTF}) includes $3,424$ videos from $1,595$ different people. All of the video sequences are from Youtube, and the average length of a video clip of YTF is about $181$ frames. The unrestricted with labeled outside data protocol contains $5,000$ video pairs, half of which belong to the same identities and the others come from different identities. Recent datasets \textbf{CPLFW}~\cite{zheng2018cross} and \textbf{CALFW}~\cite{zheng2017cross} extend LFW from the perspective of large-pose and large-age types, respectively. And \textbf{CFP-FP}~\cite{Sengupta2016Frontal} isolates pose variation with extreme poses like profile, where many features are occluded. The dataset contains $10$ frontal and $4$ profile images of $500$ individuals. Similar to LFW, the standard data protocol has defined $10$ splits, and contains $350$ pairs with the same identities and $350$ pairs with different identities.
\textbf{MegaFace}~\cite{kemelmacher2016megaface} is a challenging dataset recently released for large-scale FR, which includes 1M images of 690K different individuals for evaluation. 
\section{LGC Attack}
\subsection{More details about LGC}
\begin{figure}[t]
\centering
\includegraphics[width=0.8\linewidth]{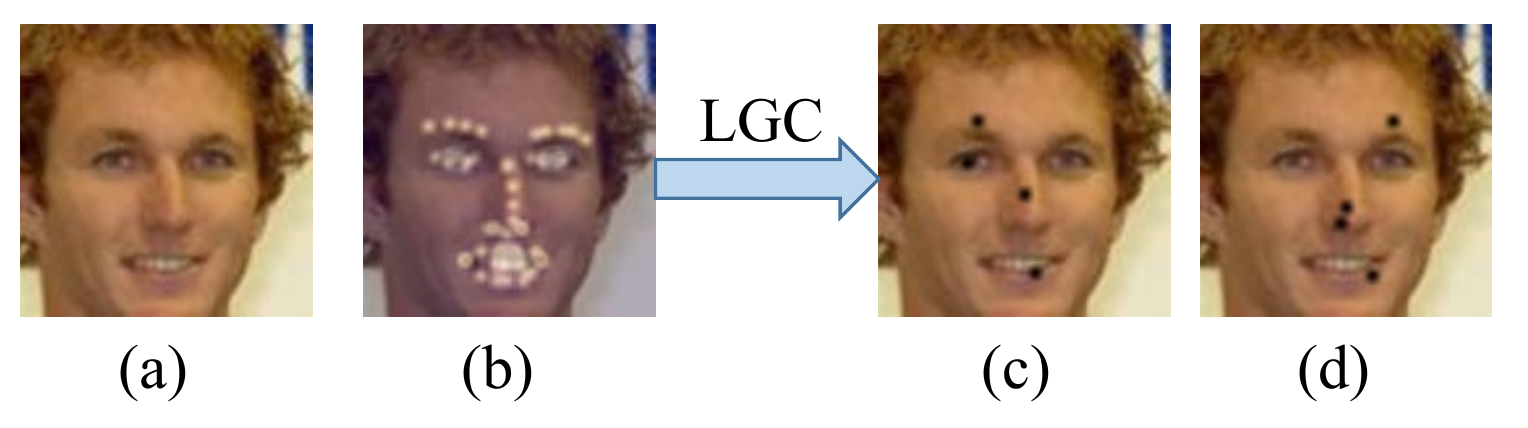}
\vspace{-1ex}
\caption{The illustration of the landmark-guided cutout method, where (a) is the original input, (b) has fixed number of landmarks. We randomly sample face landmarks in (b) as a mask in the process of generating adversarial examples, e.g., (c) and (d).}
\label{fig:lgc}
\vspace{-1ex}
\end{figure}

LGC occludes units from \emph{prominent} input regions of the images, making the network focus on less prominent regions and obtain more transferable adversarial examples. To achieve this goal, we apply a spatial prior by locating face landmarks by using face landmark detection~\cite{gutta2002investigation}, as illustrated in Fig.~\ref{fig:lgc}.  Specifically, we apply a fixed-size mask to randomly sampled $m$ locations as center points from face landmarks, and place a square patch around those locations. A simple combination of MIM and landmark-guided cutout can give rise to the landmark-guided cutout iterative method as
\begin{equation}\small
\begin{gathered}
\label{eq:a-cim}
     \bm{g}_{t+1} = \mu \cdot \bm{g}_t + \frac{\nabla_{\bm{x}}\mathcal{D}_f({M}_t\odot{\bm{x}_t^{adv}},\bm{x}^{r})}{\|\nabla_{\bm{x}}\mathcal{D}_f({M}_t\odot{\bm{x}_t^{adv}},\bm{x}^{r})\|_1}; \quad
     \bm{x}^{adv}_{t+1}=\mathrm{clip}_{\bm{x},\epsilon}(\bm{x}^{adv}_t+\alpha\cdot\mathrm{sign}(\bm{g}_{t+1})).
\end{gathered}
\end{equation}

where $M_{t} \in \{0,1\}^{d}$ is a binary mask, $\odot$ is the element-wise dot product, and $d$ is the dimension of the face image.
In the $t$-th iteration, after initializing the values of $M_{t}$ as $1$, the some randomly sampled fixed-size small square regions are set to $0$ to form $M_{t}$. The algorithm is summarized in Algorithm~\ref{algo1}.

\begin{algorithm}[t]\small
    \caption{LGC Attack}\label{algo1}
\begin{algorithmic}[1]
\Require{An input $\bm{x}$; a reference image $\bm{x}^r$; a feature representation function $f$; a face landmark detection $\mathcal{LD}$;the size of perturbation $\epsilon$; learning rate $\alpha$; iteration $N$; location numbers $m$.}
\Ensure{An adversarial image.}

\State $\bm{x}_0^{adv} = \bm{x}$; $\bm{g}_{0} = 0$; $P = \mathcal{LD}(\bm{x})$;
\For{$t = 0$ {\bfseries to} $N-1$} 
\State Initialize a mask ${M}_{t}$ with $\{1\}^{d}$;
\State Randomly sample m locations $P_{m}$ from $P$, and place a square zero-patch on $P_{m}$ to form $M_{t}$;
\State Input $M_t\odot{\bm{x}_t^{adv}}$ and obtain the gradient $\nabla_{\bm{x}}\mathcal{D}_f(M_t\odot{\bm{x}_t^{adv}},\bm{x}^{r})$; 
\State Update  $\bm{g}_{t+1}$ and $\bm{x}_{t+1}^{adv}$ by Eq.~\eqref{eq:a-cim}.
\EndFor
\end{algorithmic}
\end{algorithm}

\subsection{Comparison of CIM and LGC}
\begin{figure*}[t]
\begin{center}
\includegraphics[width=0.99\linewidth]{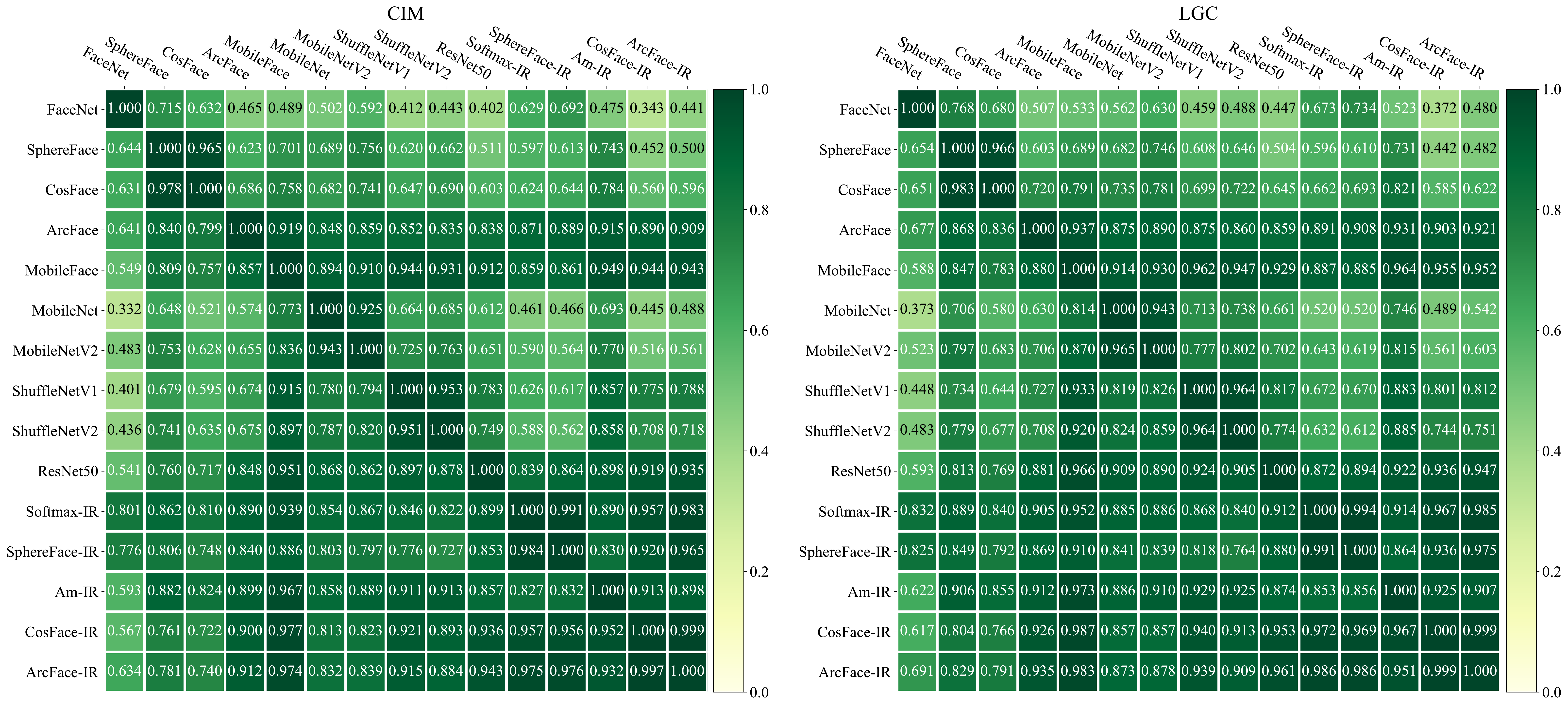}
\end{center}
\vspace{-2ex}
\caption{The success rates of the $15$ models against black-box dodging attacks based on CIM and LGC under the $\ell_{\infty}$ norm.}
\label{fig:black-d-linf-cim}
\vspace{-1ex}
\begin{center}
\includegraphics[width=0.99\linewidth]{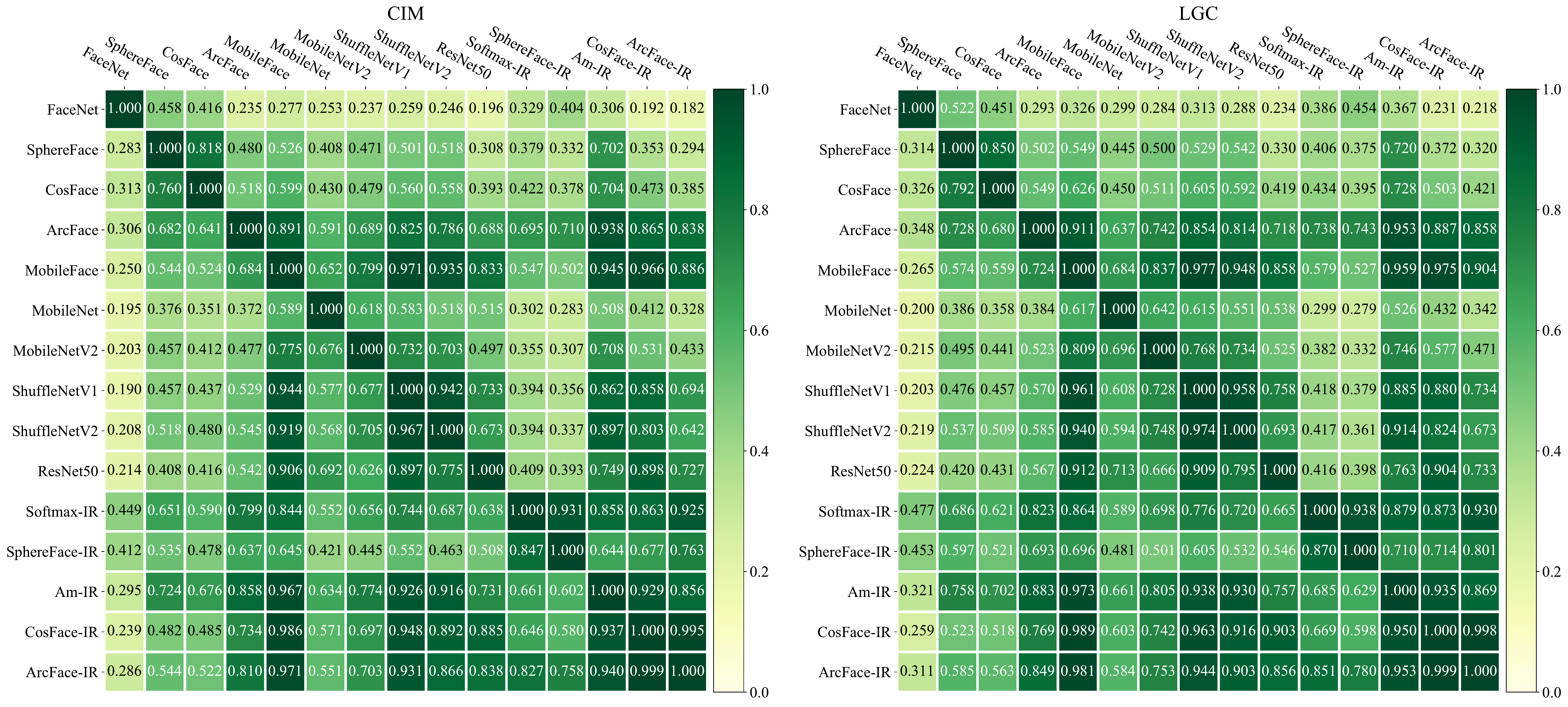}
\end{center}
\vspace{-2ex}
\caption{The success rates of the $15$ models against black-box impersonation attacks based on CIM and LGC under the $\ell_{\infty}$ norm.}
\label{fig:black-p-linf-cim}
\vspace{-1ex}
\end{figure*}

We compare the performance of Cutout (CIM) and Landmark-Guided Cutout (LGC) on the LFW dataset.
For CIM, we apply a default square zero-mask with a side length of 10~\cite{devries2017improved} to
a random location at each iteration. As for LGC, the mask consists of
four small squares, making a randomly selected key face landmark as the center. Note that we ensure
that the total mask area used for CIM and LGC is equal. Fig.~\ref{fig:black-d-linf-cim} and Fig.~\ref{fig:black-p-linf-cim} show the success rates of the $15$ models against black-box dodging and impersonation attacks based on CIM and LGC under the $\ell_{\infty}$ norm. LGC achieves more performance under the dodging and impersonation setting. The results also demonstrate that
the  adversarial examples generated by LGC are less sensitive to the discriminative regions,
thus fooling other black-box face recognition models.

\subsection{Different Quantities and Sizes for LGC}
\begin{figure*}[!htp]
\centering
\includegraphics[width=0.99\linewidth]{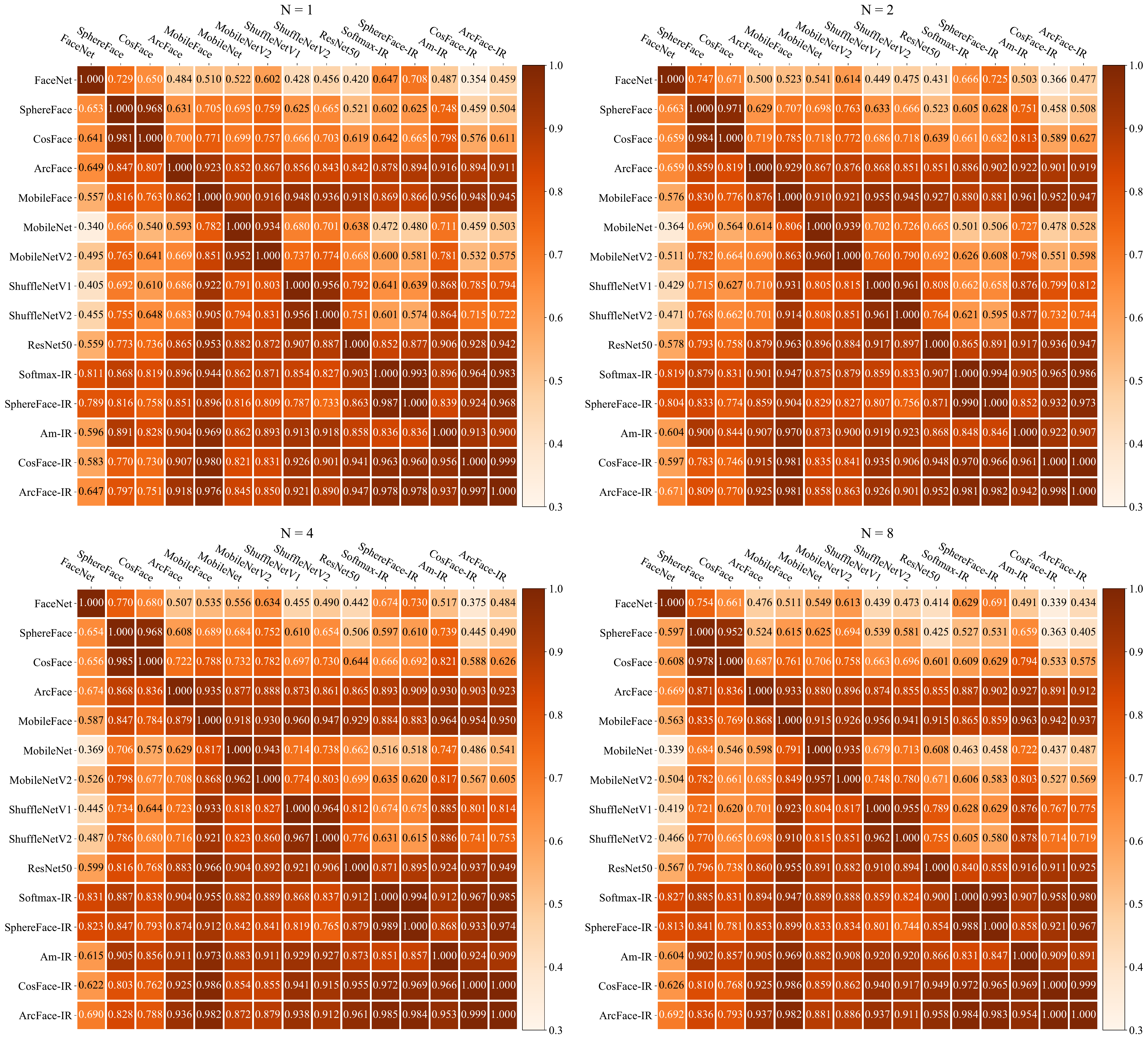}
\vspace{-2ex}
\caption{The success rates of the $15$ models against black-box doding attacks under different \emph{quantities} of black-squares for LGC under the $\ell_{\infty}$ norm on LFW.}

\label{fig:num-heatmap_d}
\vspace{-3ex}
\end{figure*}

\begin{figure*}[!htp]
\centering
\includegraphics[width=0.99\linewidth]{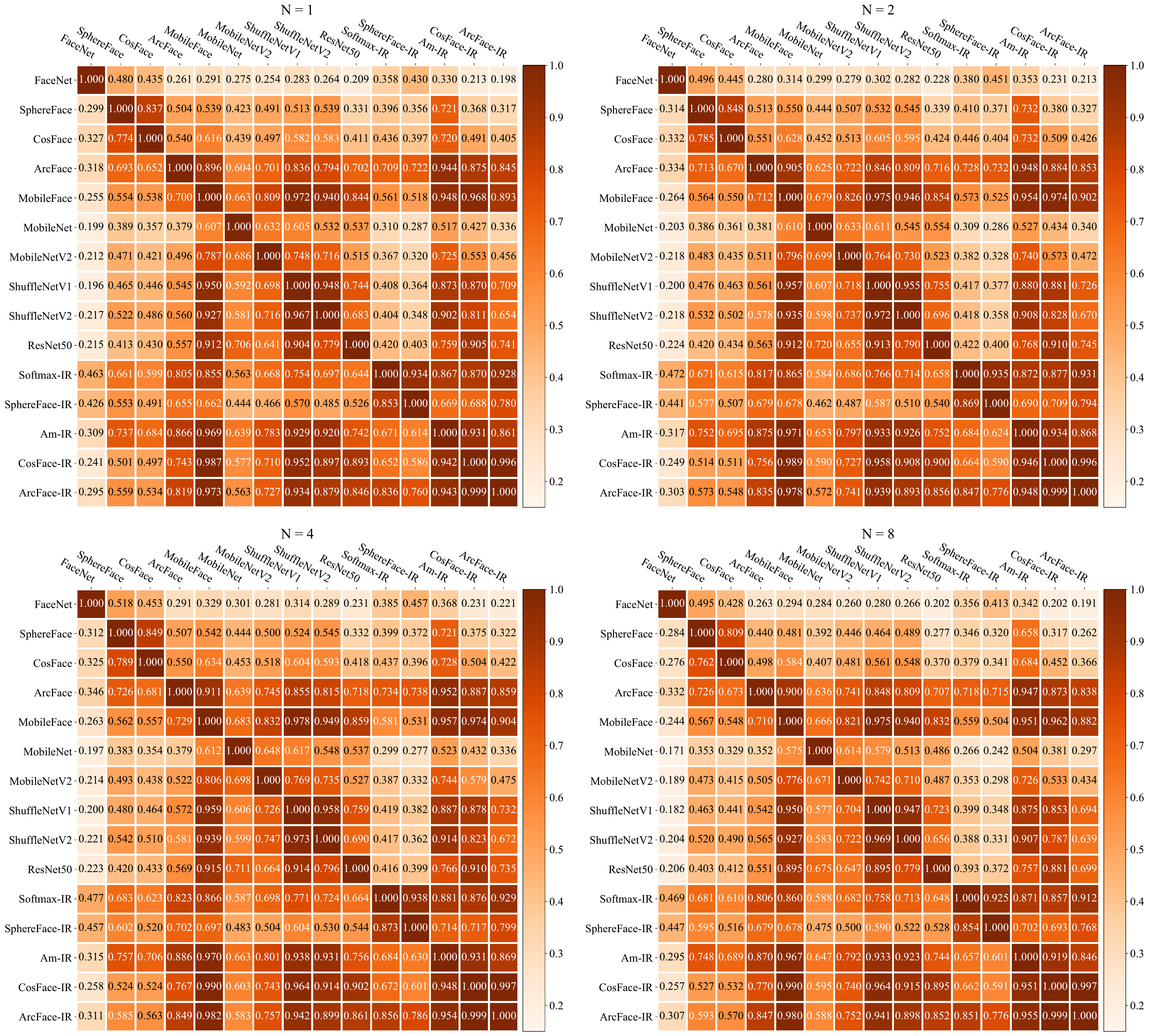}
\vspace{-2ex}
\caption{The success rates of the $15$ models against black-box impersonation attacks under different \emph{quantities} of black-squares for LGC under the $\ell_{\infty}$ norm on LFW.}
\label{fig:num-heatmap_p}
\vspace{-3ex}
\end{figure*}

\begin{figure*}[!htp]
\centering
\includegraphics[width=0.99\linewidth]{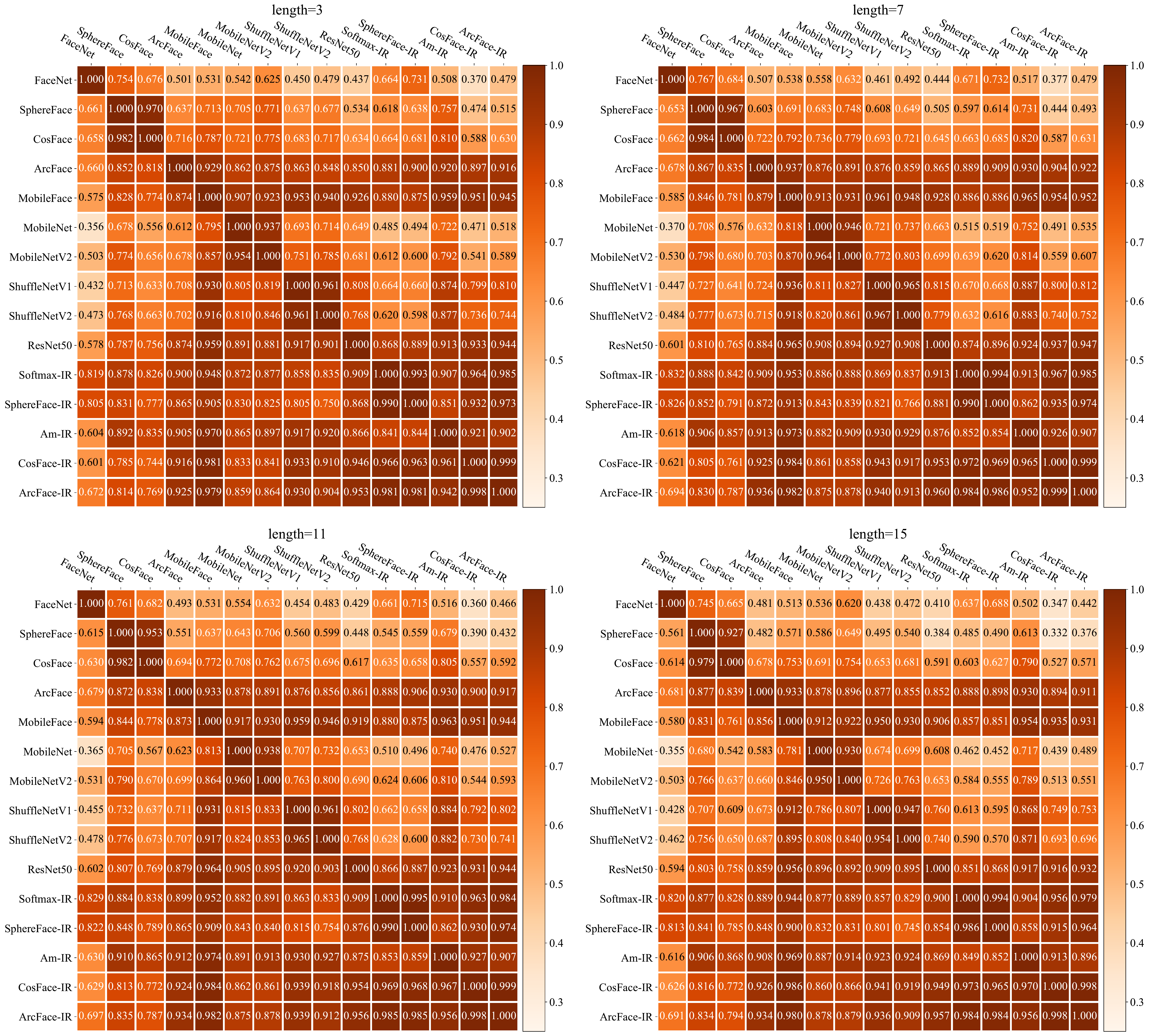}
\vspace{-2ex}
\caption{The success rates of the $15$ models against black-box doding attacks under different \emph{sizes} of black-squares for LGC under the $\ell_{\infty}$ norm on LFW.}
\label{fig:len-heatmap_d}
\vspace{-1ex}
\end{figure*}

\begin{figure*}[!htp]
\centering
\includegraphics[width=0.99\linewidth]{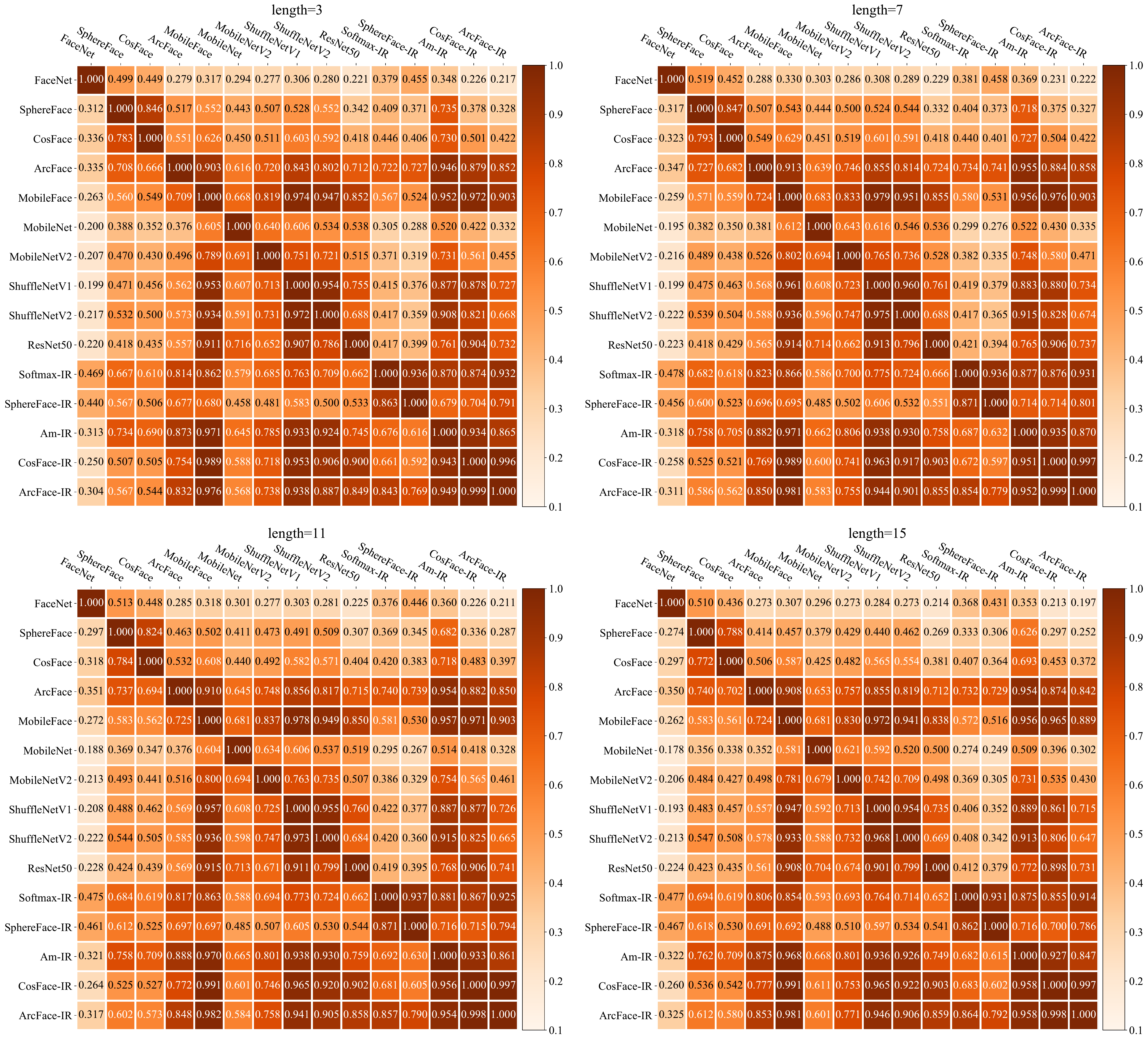}
\vspace{-2ex}
\caption{The success rates of the $15$ models against black-box impersonation attacks under different \emph{sizes} of black-squares for LGC under the $\ell_{\infty}$ norm on LFW.}
\label{fig:len-heatmap_p}
\vspace{-1ex}
\end{figure*}

In this section, we discuss the impact of different quantities and sizes on LGC. We choose a model from 15 models as the substitute model to attack the others.  Fig.~\ref{fig:num-heatmap_d} and Fig.~\ref{fig:num-heatmap_p} show the success rates of different models against black-box dodging and impersonation attacks under different \emph{quantities} of black-squares for LGC. Fig.~\ref{fig:len-heatmap_d} and Fig.~\ref{fig:len-heatmap_p} show the success rates of different models against black-box dodging and impersonation attacks under different \emph{sizes} of black-squares for LGC. 
The value in the i-th row and the j-th column of each heatmap matrix implies the attack success rate for the target model j on the adversarial samples generated by the source models i. 
We found that the best results can be achieved only with
the appropriate quantities and sizes of occlusions. Too big or
too small is not optimal.

\begin{figure*}[!htp]
\centering
\includegraphics[width=0.99\linewidth]{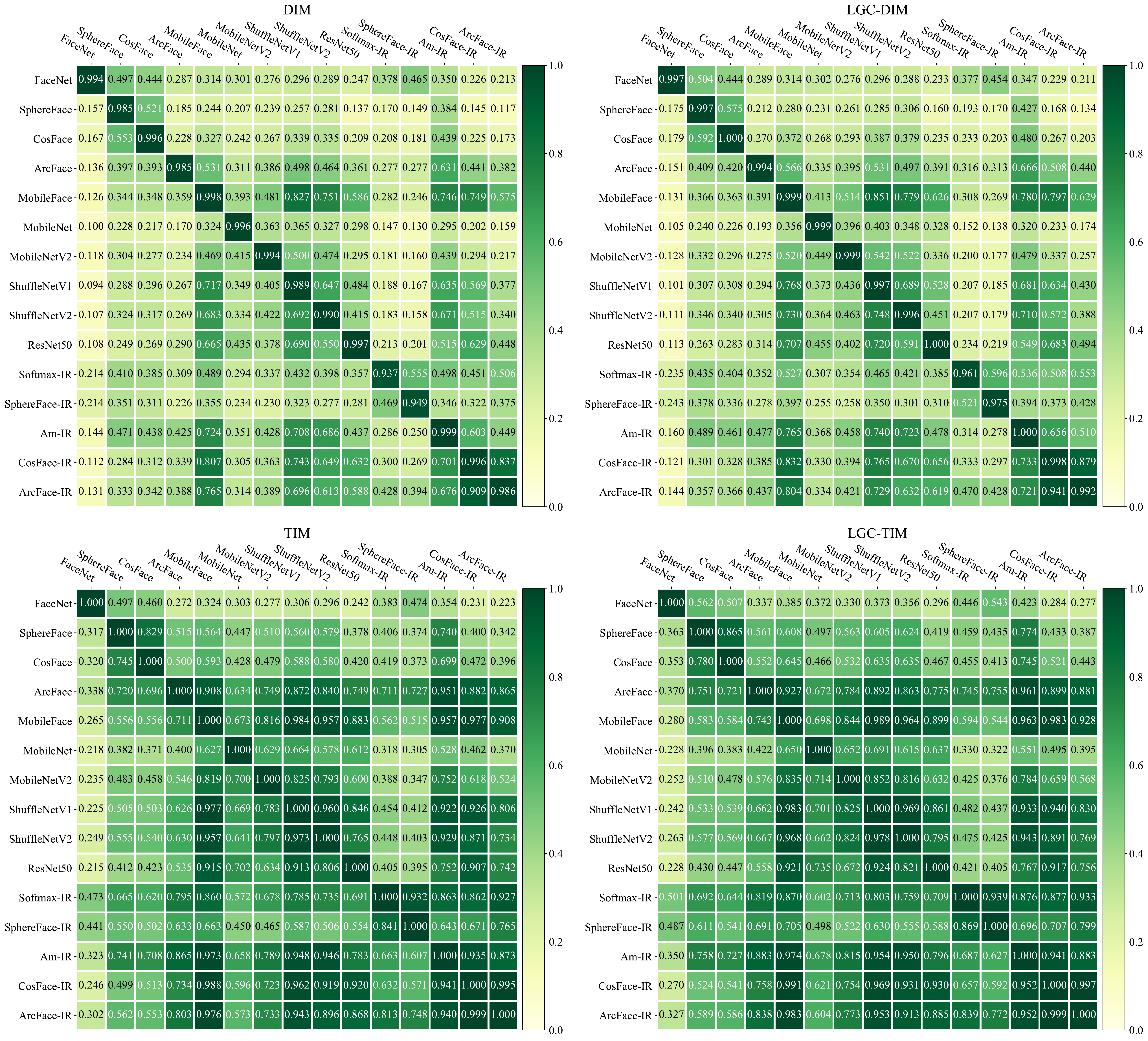}
\vspace{-2ex}
\caption{The success rates of the $15$ models against black-box impersonation attacks under the $\ell_{\infty}$ norm on LFW.}
\label{fig:exLGC}
\vspace{-2ex}
\end{figure*}

\subsection{Combinations of LGC and Other Attacks}

\textbf{DIM}~\cite{xie2019improving} relies on a stochastic transformation function to craft adversarial examples, which can be represented as
\begin{equation}
\label{eq:dim}
    \bm{x}_{t+1}^{{adv}} = \mathrm{clip}_{\bm{x},\epsilon} \big(\bm{x}_t^{{adv}} + \alpha\cdot\mathrm{sign}(\nabla_{\bm{x}}\mathcal{D}_f(T(\bm{x}_t^{adv}; p),\bm{x}^{r}))\big),
\end{equation}
where $T(\bm{x}_t^{{adv}}; p)$ refers to some transformation to diversify the input with probability $p$. Therefore, a simple combination of LGC and DIM named \textbf{LGC-DIM} can be denoted as  

\begin{equation}
\label{eq:dim-lgc}
    \bm{x}_{t+1}^{{adv}} = \mathrm{clip}_{\bm{x},\epsilon} \big(\bm{x}_t^{{adv}} + \alpha\cdot\mathrm{sign}(\nabla_{\bm{x}}\mathcal{D}_f(T(\mathrm{M}_t\odot{\bm{x}_t^{adv}}; p),\bm{x}^{r}))\big),
\end{equation}
where $M_{t} \in \{0,1\}^{d}$ is a binary mask and $\odot$ is the element-wise dot product, and $d$ is the dimension of the face image.
In the $t$-th iteration, some randomly sampled fixed-size small square regions are set to $0$ to form $M_{t}$.

\textbf{TIM}~\cite{dong2019evading} integrates the translation-invariant method into BIM by convolving the gradient with the pre-defined kernel $\bm{W}$ as
\begin{equation}
\label{eq:tim}
    \bm{x}_{t+1}^{{adv}} = \mathrm{clip}_{\bm{x},\epsilon} \big(\bm{x}_t^{{adv}} + \alpha\cdot\mathrm{sign}(\bm{W} * \nabla_{\bm{x}}\mathcal{D}_f(\bm{x}_t^{adv},\bm{x}^{r}))\big).
\end{equation}
Similarly, a simple combination of LGC and TIM named \textbf{LGC-TIM} can be denoted as
\begin{equation}
\label{eq:tim-lgc}
    \bm{x}_{t+1}^{{adv}} = \mathrm{clip}_{\bm{x},\epsilon} \big(\bm{x}_t^{{adv}} + \alpha\cdot\mathrm{sign}(\bm{W} * \nabla_{\bm{x}}\mathcal{D}_f(\mathrm{M}_t\odot{\bm{x}_t^{adv}},\bm{x}^{r}))\big).
\end{equation}

Fig.~\ref{fig:exLGC} show the success rates of the 15 models against different black-box impersonation attacks based on DIM, LGC-DIM, TIM and LGC-TIM under the $\ell_{\infty}$ norm.  After incorporating LGC into DIM and TIM, attack algorithms achieve better performance, which also demonstrate the consistent effectiveness for the adversarial examples generated by LGC in terms of FR.  Therefore, LGC can be flexibly applied to any black-box attack method to improve the performance.

%
%

\begin{figure*}[t]
\centering
\includegraphics[width=0.95\linewidth]{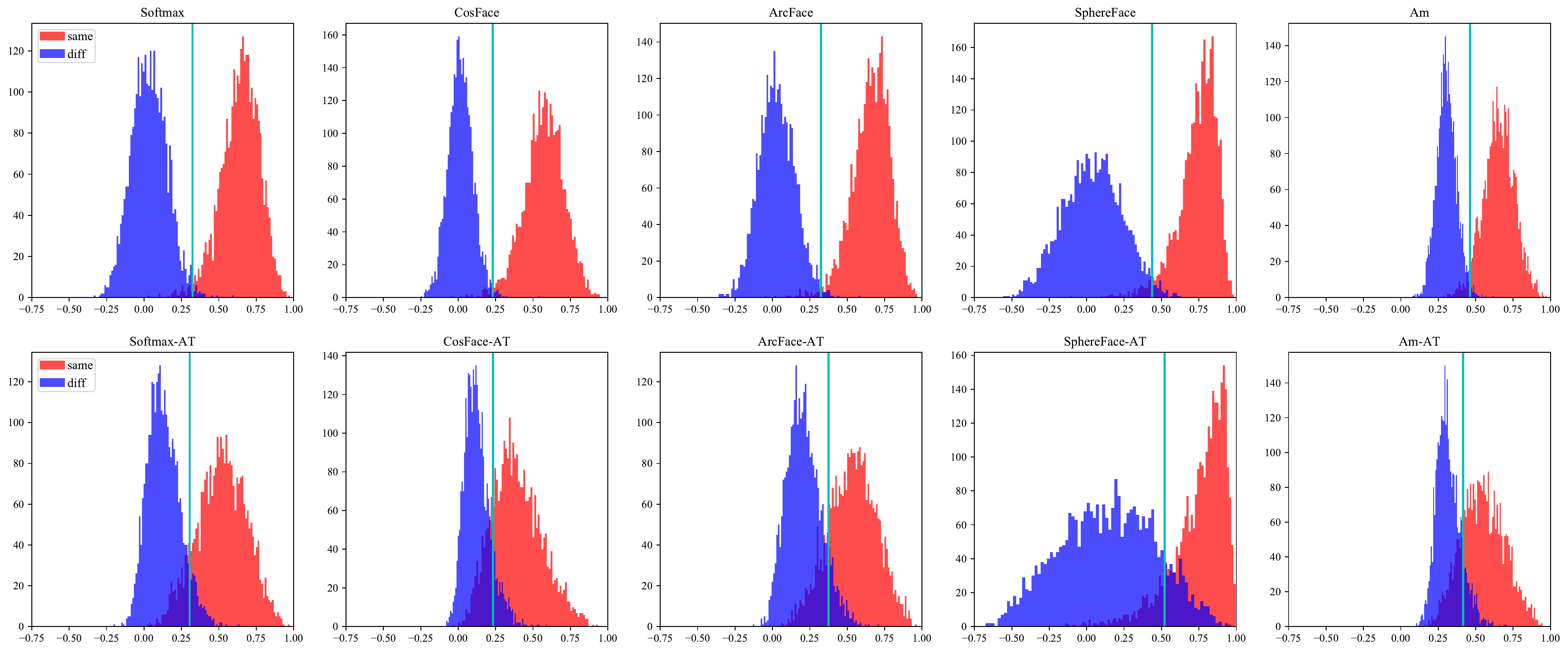}
\vspace{-2ex}
\caption{Distance distributions of both same and different pairs on LFW under normal training and adversarial training. The cyan lines refer to the thresholds.}
\label{fig:similarity}
\vspace{-1ex}
\end{figure*}

\section{Defense details}

we mainly consider the representative AT framework named PGD-AT~\cite{madry2018towards}, and adaptively integrate different loss functions in Tab.~\ref{tab:loss} an AT procedure. And Fig.~\ref{fig:similarity} also shows the distance distribution between normal and adversarial training.

\begin{table*}[!htp]
    \footnotesize
    \setlength{\tabcolsep}{9pt}
  \caption{Formulations of different loss functions $\mathcal{L}(\mathbf{z}_{i}, y_{i}, \mathbf{W})$. $\mathbf{z}_{i} = f(\bm{x}_{i})$ is the extracted feature vector with the input $\bm{x}_{i}$, and $\theta_{i, j}$ indicates the angle between $\mathbf{W}_{i}$ and $\mathbf{z}_{j}$. Note that Am-Softmax~\cite{wang2018additive} and LMCL~\cite{wang2018cosface} possess the same formulation of loss.}
  \label{tab:loss}
  \begin{center}
  \begin{tabular}{c|c}
\hline
\!\!Method\! &  Loss Function  $\mathcal{L}$ \\

\hline
Softmax & $\frac{1}{N} \sum_{i}-\log \left(\frac{\mathbf{W}_{y_{i}} \cdot \mathbf{z}_{i}}{\sum_{k} e^{\mathbf{W}_{k} \mathbf{z}_{i}}}\right)$ \\

\hline 
A-Softmax~\cite{liu2017sphereface} & $\frac{1}{N} \sum_{i}-\log \left(\frac{e^{\left\|\mathbf{W}_{y_{i}}\right\| \cdot\left\|\mathbf{z}_{i}\right\| \cos \left(m \cdot \theta_{y_{i}, i}\right)}} {e^{\|\mathbf{W}_{y_{i}}\| \cdot\|\mathbf{z}_{i}\| \cdot \cos \left(m \cdot \theta_{y_{i}, i}\right)}+\sum_{k \neq y_{i}} e^{\|\mathbf{W}_{k}\| \cdot \|\mathbf{z}_{i}\| \cdot \cos \theta_{k, i}}}\right)$ \\
\hline
Am-Softmax~\cite{wang2018additive} & $\frac{1}{N} \sum_{i}-\log \left(\frac{e^{s \cdot\left(\cos \theta_{y_i, i}-m\right)}}{e^{s\left(\cos \theta_{y_{i}, t}-m\right)}+\sum_{k \neq y_{i}} e^{s \cos \theta_{k, i}}}\right)$ \\
\hline
LMCL~\cite{wang2018cosface} & $\frac{1}{N} \sum_{i}-\log \left(\frac{e^{s \cdot\left(\cos \theta_{y_i, i}-m\right)}}{e^{s\left(\cos \theta_{y_{i}, t}-m\right)}+\sum_{k \neq y_{i}} e^{s \cos \theta_{k, i}}}\right)$ \\

\hline
ArcFace~\cite{deng2018arcface} & $\frac{1}{N} \sum_{i}-\log \left(\frac{e^{s \cdot \cos \left(\theta_{y_{i}, i}+m\right)}}{e^{s \cdot \cos \left(\theta_{y_{i}, i}+m\right)}+\sum_{k \neq y_{i}} e^{s \cdot \cos \theta_{k, i}}}\right)$ \\
\hline
   \end{tabular}
  \end{center}
  
\end{table*}

\section{Evaluation on Commercial API}
\begin{figure*}[t]
\centering
\includegraphics[width=0.99\linewidth]{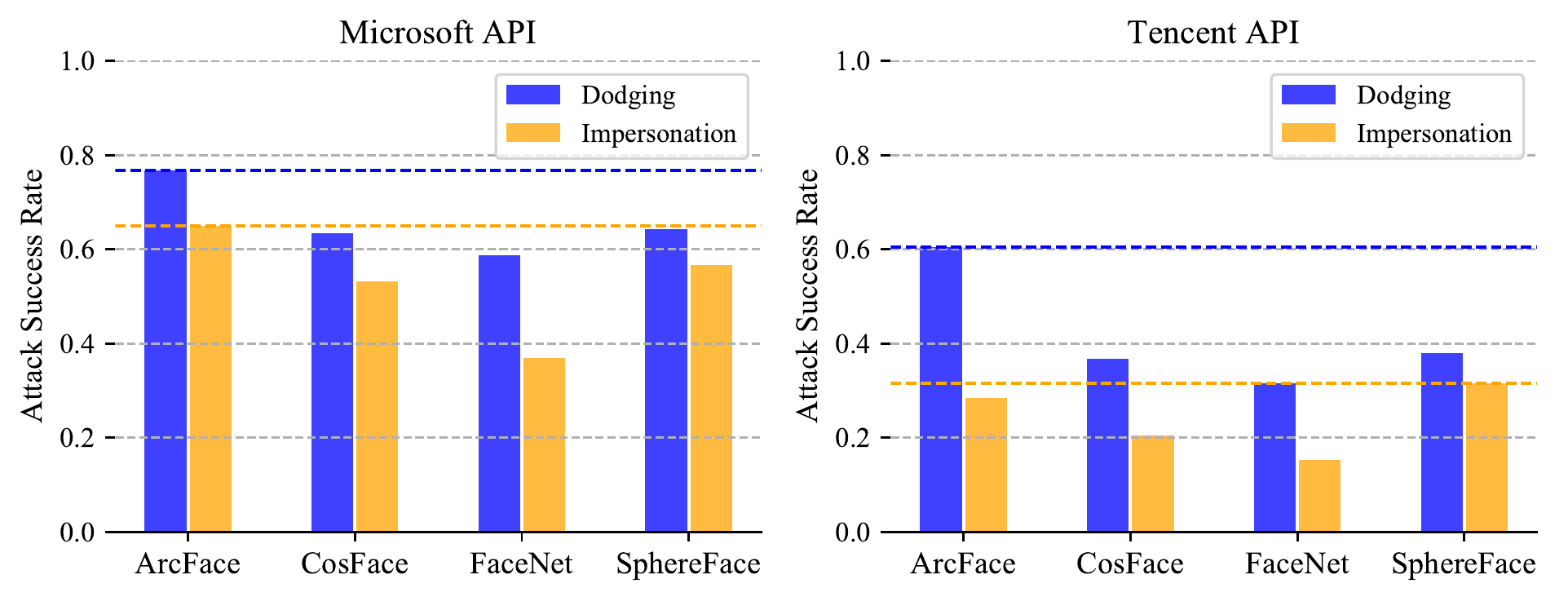}
\vspace{-2ex}
\caption{The success rates of commercial APIs against black-box dodging and impersonation attacks under the $\ell_{\infty}$ norm on LFW.}
\label{fig:api}
\vspace{-1ex}
\end{figure*}

We evaluate adversarial robustness on 2 commercial FR API services available at Microsoft API~\footnote{\url{https://docs.microsoft.com/}} and Tencent API~\footnote{\url{https://cloud.tencent.com/}}. The
working mechanism and training data are completely unknown
for us. We follow their original threshold ranges, and compute the optimal thresholds on LFW for dodging and impersonation attacks. Thus after  crafting adversarial examples based on proposed LCG against ArcFace, CosFace, FaceNet and SphereFace, we then feed the generated images into black-box APIs for benchmarking adversarial robustness. Fig.~\ref{fig:api} shows the success rates of commercial APIs against black-box dodging and impersonation attacks under the $\ell_{\infty}$ norm on LFW. Microsoft API obtains unsatisfying performance on robustness with an attack success rate of $76.7\%$ for dodging attack, $65.0\%$ for impersonation attack. In contrast, although Tencent API presents progressive resistant performance against adversarial examples, it does not exhibit reliable effectiveness on robustness evaluation for security-critical FR systems. Thus targeted exploration of designing models is significantly taking into account robustness in the future.

\section{Full Evaluation Results}
In addition to the LFW dataset mentioned, we also perform an adversarial evaluation on YouTube Faces Database
(YTF)~\cite{wolf2011face}. Besides, experiments on larger and more challenging face verification datasets, e.g., CFP-FP~\cite{Sengupta2016Frontal}, are shown in the following section.
\subsection{Additional results on the LFW dataset}
\begin{figure*}[t]
\begin{center}
\includegraphics[width=0.8\linewidth]{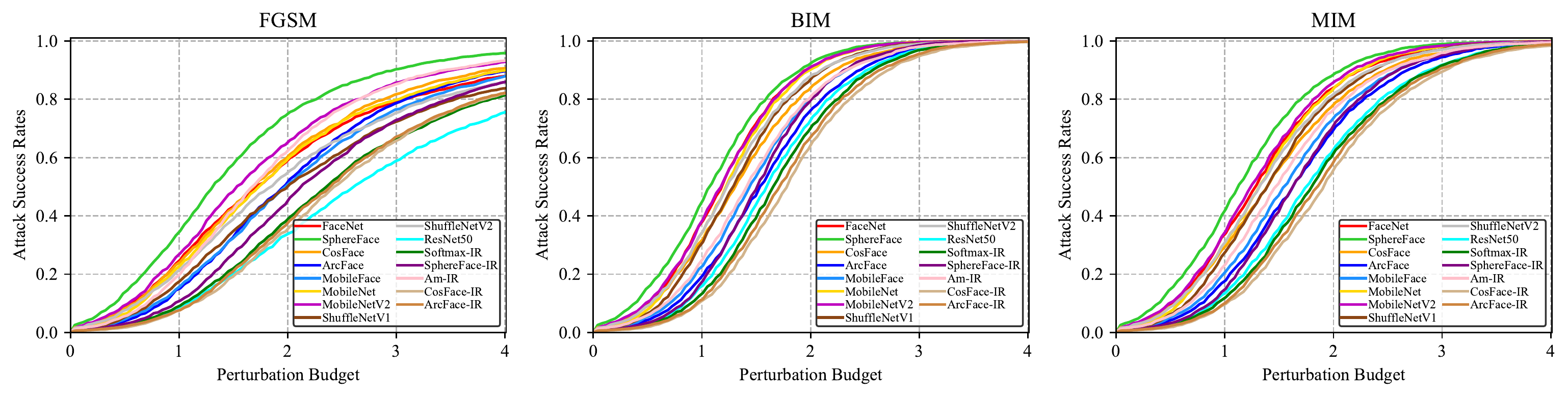}
\end{center}
\vspace{-2ex}
\caption{\textit{Asr vs. perturbation budget} curves of the $15$ models against dodging attacks under the $\ell_{\infty}$ norm.}
\label{fig:white-d-linf-asr-pert}
\begin{center}
\includegraphics[width=0.8\linewidth]{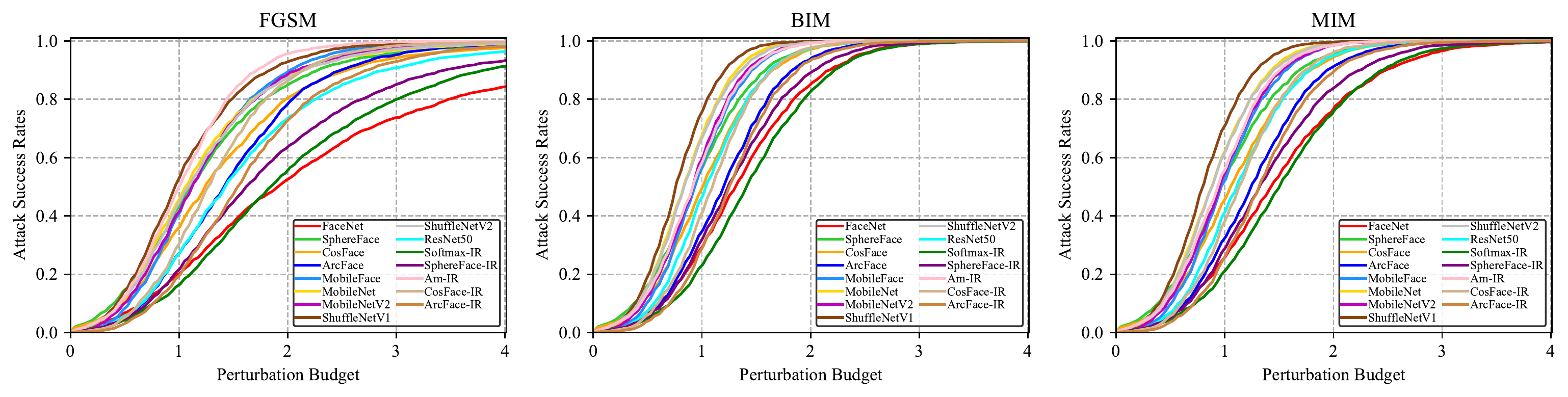}
\end{center}
\vspace{-3ex}
\caption{\textit{Asr vs. perturbation budget} curves of the $15$ models against impersonation attacks under the $\ell_{\infty}$ norm.}
\label{fig:white-p-linf-asr-pert}
\vspace{-2ex}
\end{figure*}

We present the curves of FGSM, BIM, MIM under the $\ell_\infty$ norm in  Fig.~\ref{fig:white-d-linf-asr-pert} and Fig.~\ref{fig:white-p-linf-asr-pert}. 

\subsection{Evaluation on the YTF dataset}
YTF contains 5000 video pairs, half of
which belong to the same identities and others come from different identities. We select an intermediate frame as the representative image to perform our task.  We perform dodging attacks based on the pairs of images with the same identities, and impersonation
attacks based on the pairs with different identities. The parameter setting is the same as LFW.

\begin{figure*}[!htp]
\begin{center}
\includegraphics[width=0.9\linewidth]{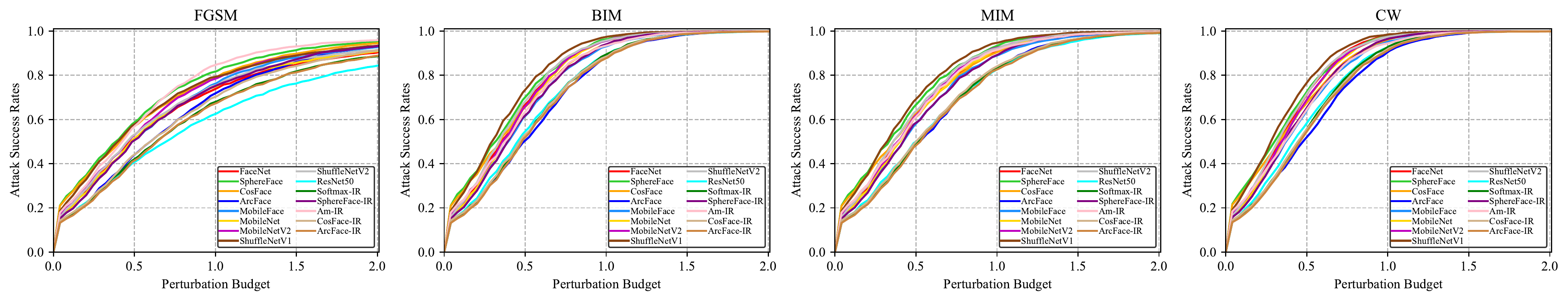}
\end{center}
\vspace{-2ex}
\caption{The \textit{attack success rate vs. perturbation budget} curves of the $15$ models against dodging attacks under the $\ell_2$ norm on YTF.}
\label{fig:ytf-white-d-l2-asr-pert}
\begin{center}
\includegraphics[width=0.9\linewidth]{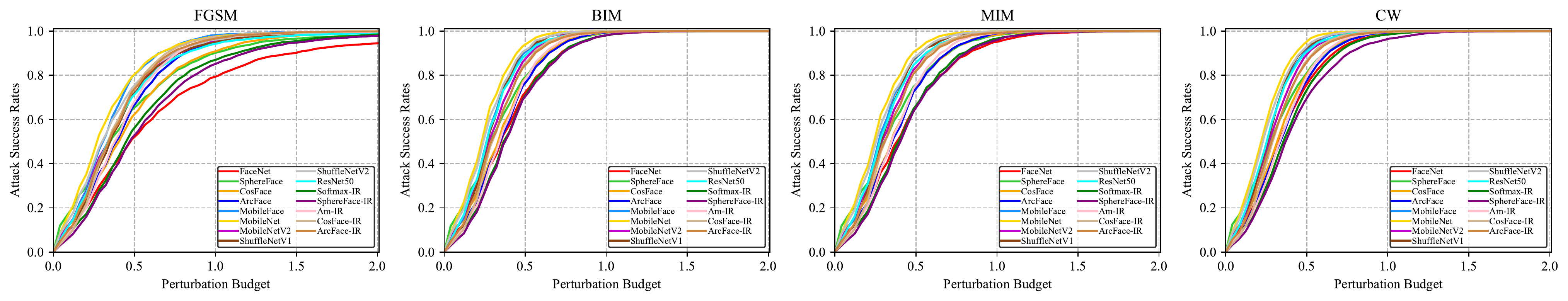}
\end{center}
\vspace{-2ex}
\caption{The \textit{attack success rate vs. perturbation budget} curves of the $15$ models against impersonation attacks under the $\ell_2$ norm on YTF.}
\label{fig:ytf-white-p-l2-asr-pert}
\vspace{-2ex}
\end{figure*}

\begin{figure*}[!htp]
\begin{center}
\includegraphics[width=0.85\linewidth]{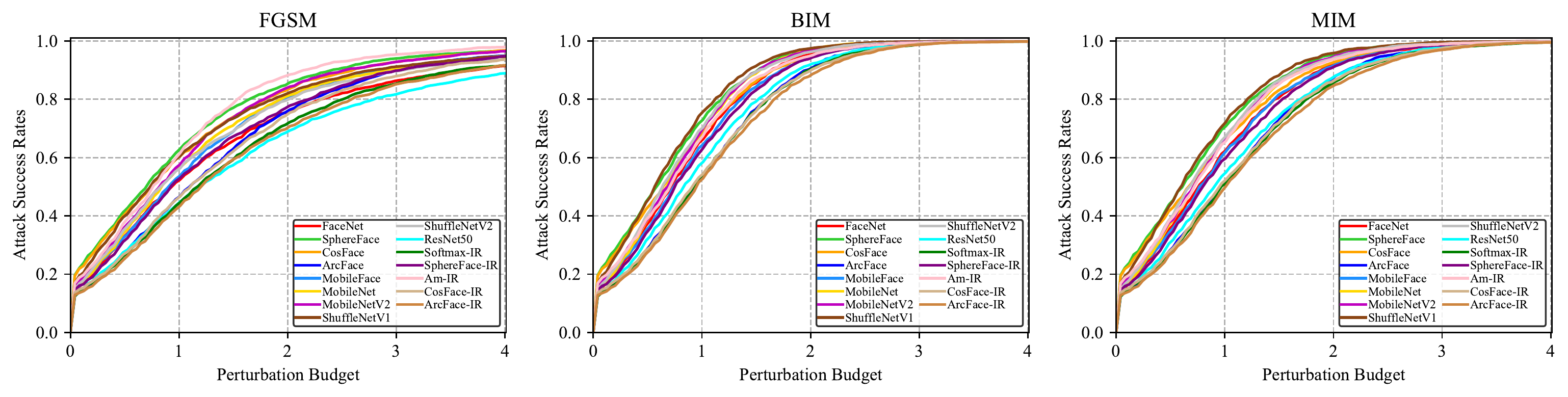}
\end{center}
\vspace{-2ex}
\caption{The \textit{attack success rate vs. perturbation budget} curves of the $15$ models against dodging attacks under the $\ell_{\infty}$ norm on YTF.}
\label{fig:ytf-white-d-linf-asr-pert}
\begin{center}
\includegraphics[width=0.85\linewidth]{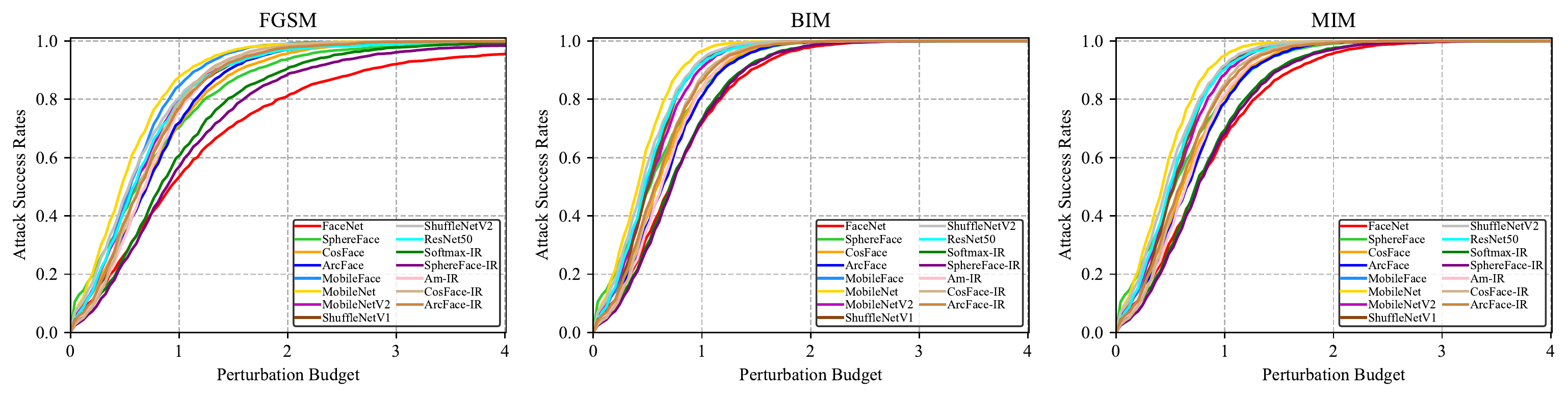}
\end{center}
\vspace{-2ex}
\caption{The \textit{attack success rate vs. perturbation budget} curves of the $15$ models against impersonation attacks under the $\ell_{\infty}$ norm on YTF.}
\label{fig:ytf-white-p-linf-asr-pert}
\vspace{-2ex}
\end{figure*}

\textbf{White-box attacks:} We show some robustness curves for white-box attacks and provide attack success rate vs. perturbation
budget of face recognition models against dodging
and impersonation attacks under the $\ell_{2}$ and $\ell_{\infty}$ norm. Fig.~\ref{fig:ytf-white-d-l2-asr-pert} and  Fig.~\ref{fig:ytf-white-p-l2-asr-pert} show the \emph{attack success rate vs. perturbation budget} curves of the 15 models against dodging and impersonation attacks under the $\ell_{2}$ norm. Fig.~\ref{fig:ytf-white-d-linf-asr-pert} and  Fig.~\ref{fig:ytf-white-p-linf-asr-pert} show the \emph{attack success rate vs. perturbation budget} curves of the 15 models against dodging and impersonation attacks under the $\ell_{\infty}$ norm.

\begin{figure*}[!htp]
\centering
\includegraphics[width=0.99\linewidth]{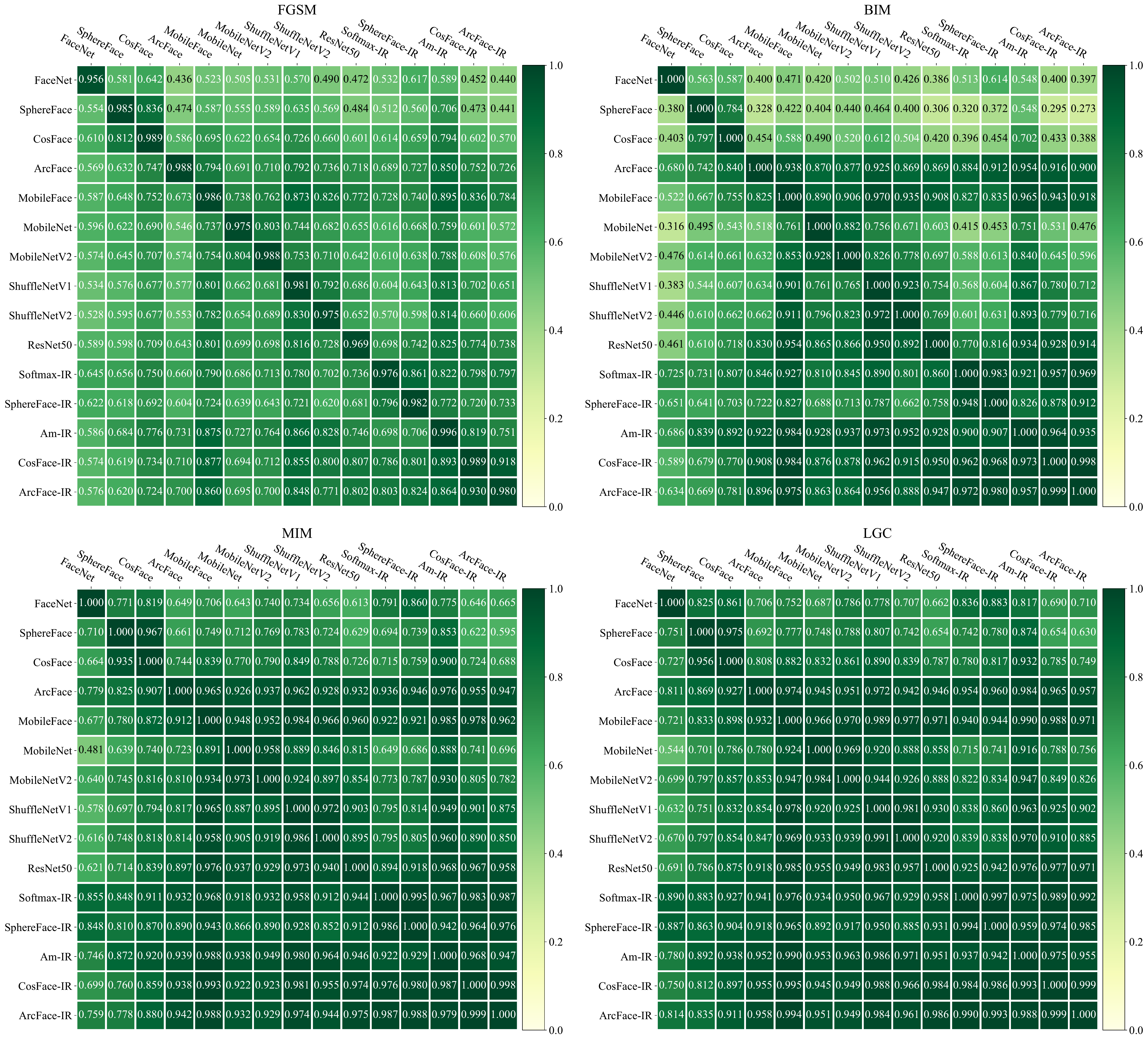}
\vspace{-2ex}
\caption{The success rates of the $15$ models against black-box dodging attacks under the $\ell_{\infty}$ norm on YTF.}
\label{fig:ytf-heatmap_d}
\vspace{-1ex}
\end{figure*}

\begin{figure*}[!htp]
\centering
\includegraphics[width=0.99\linewidth]{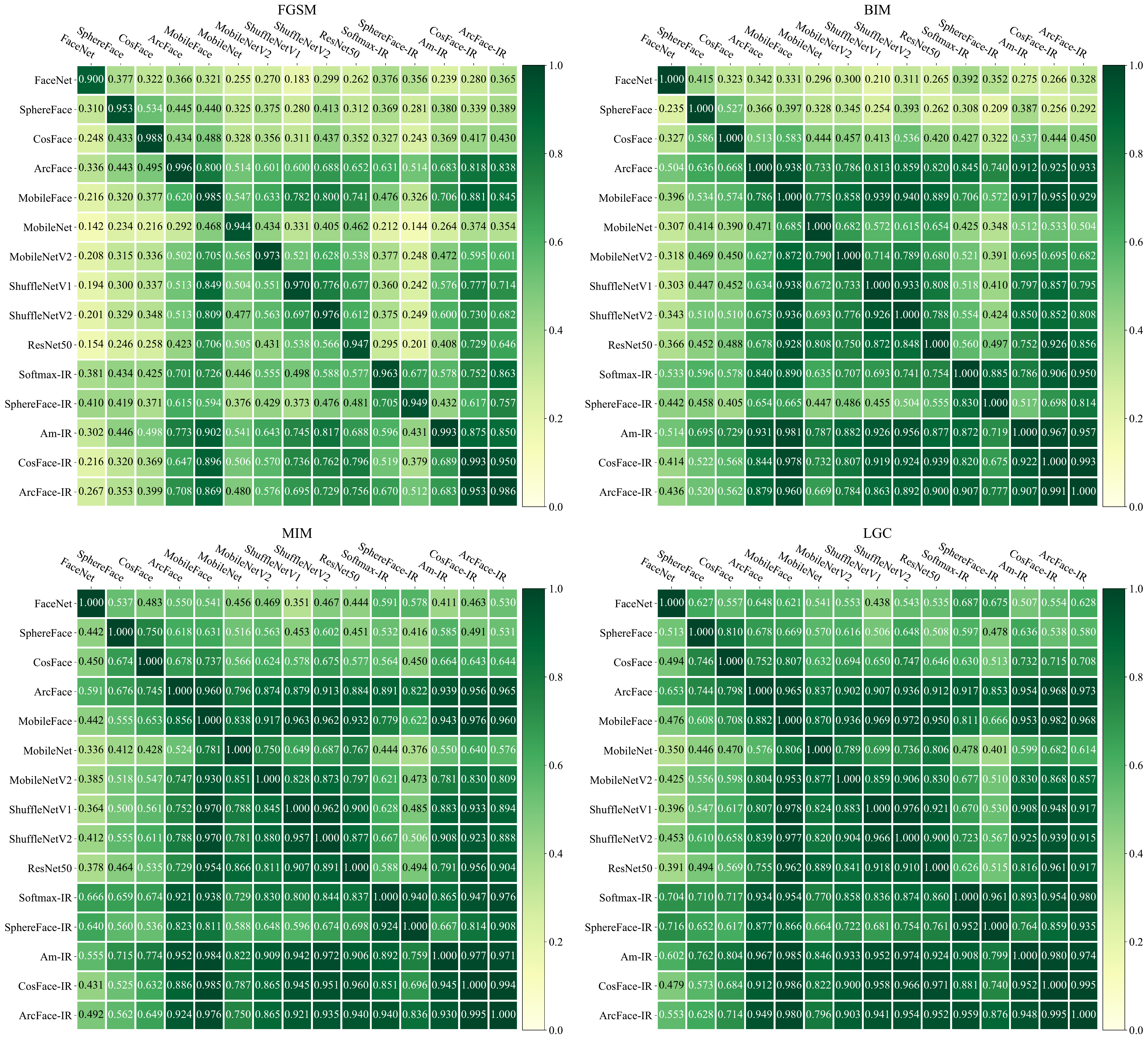}
\vspace{-2ex}
\caption{The success rates of the $15$ models against black-box impersonation attacks under the $\ell_{\infty}$ norm on YTF.}
\label{fig:ytf-heatmap_p}
\end{figure*}

\textbf{Transfer-based black-box attacks:} We show the attack success rate of the 15 models against dodging
and impersonation attacks based on black-box FGSM,
BIM, MIM, and LGC methods under the $\ell_{\infty}$ norm in Fig.~\ref{fig:ytf-heatmap_d} and Fig.~\ref{fig:ytf-heatmap_p}.

\subsection{Evaluation on the CFP-FP dataset}

CFP-FP dataset isolates pose variation with extreme poses like profile, where many features are occluded. The dataset contains $10$ frontal and $4$ profile images of $500$ individuals. Similar to LFW, the standard data protocol has defined $10$ splits, and contains $350$ pairs with the same identities and $350$ pairs with different identities.

\begin{figure*}[!htp]
\begin{center}
\includegraphics[width=0.9\linewidth]{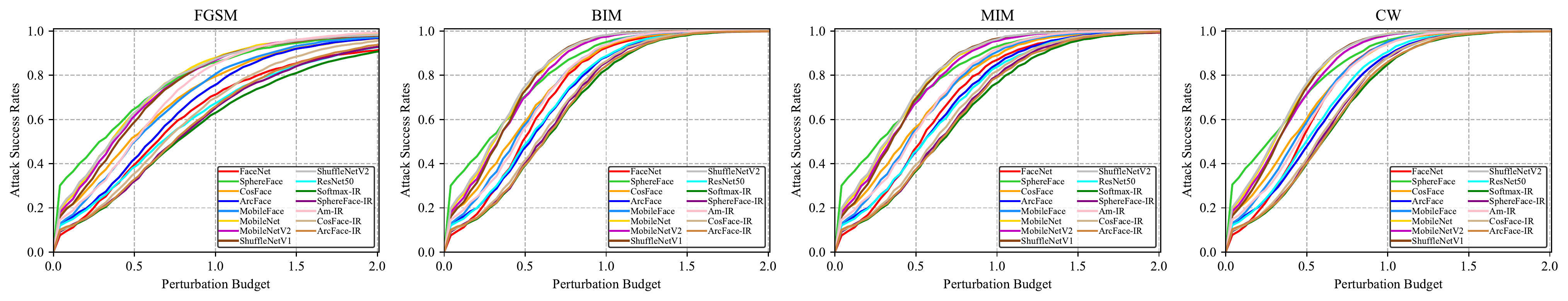}
\end{center}
\vspace{-2ex}
\caption{The \textit{attack success rate vs. perturbation budget} curves of the $15$ models against dodging attacks under the $\ell_2$ norm on CFP-FP.}
\label{fig:cfp-white-d-l2-asr-pert}
\begin{center}
\includegraphics[width=0.9\linewidth]{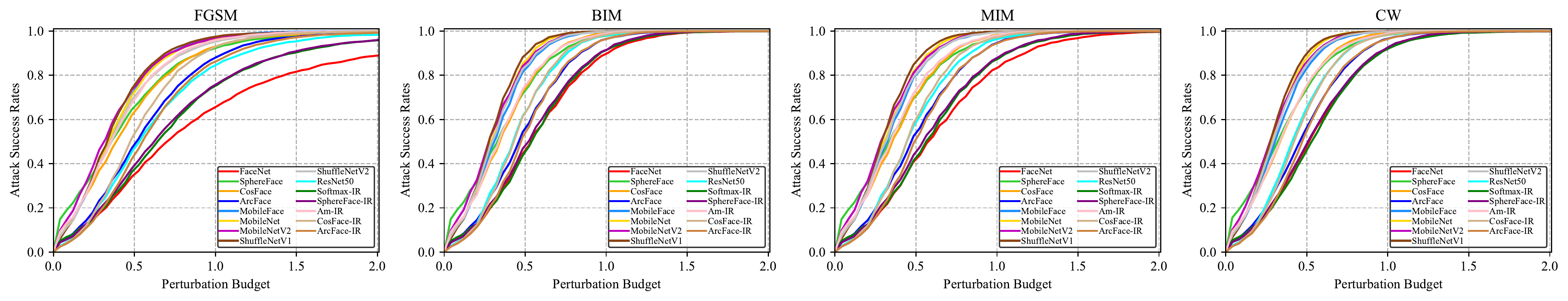}
\end{center}
\vspace{-2ex}
\caption{The \textit{attack success rate vs. perturbation budget} curves of the $15$ models against impersonation attacks under the $\ell_2$ norm on CFP-FP.}
\label{fig:cfp-white-p-l2-asr-pert}
\vspace{-2ex}
\end{figure*}

\begin{figure*}[!htp]
\begin{center}
\includegraphics[width=0.85\linewidth]{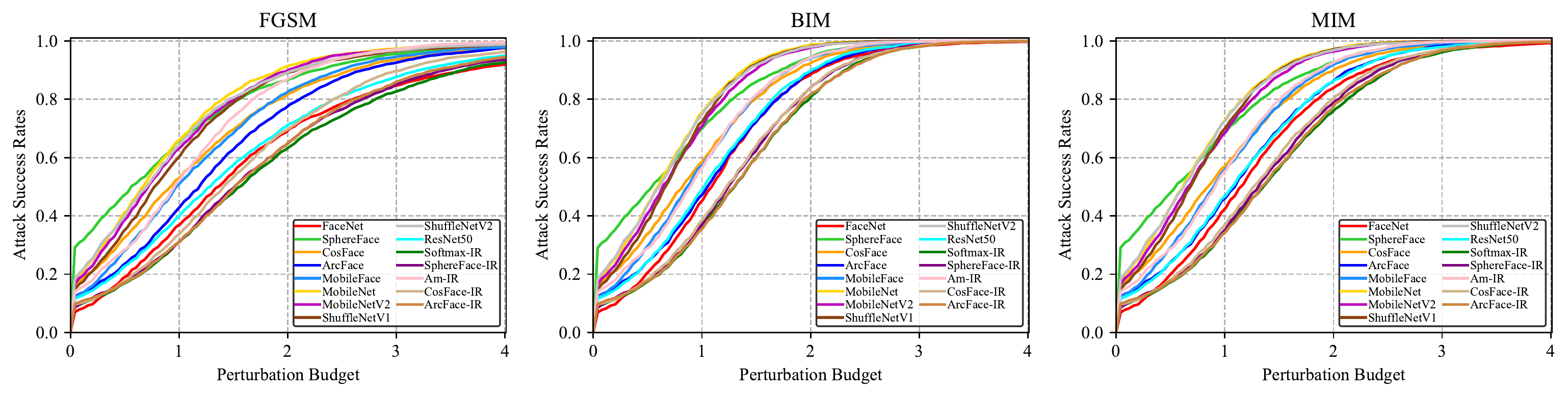}
\end{center}
\vspace{-2ex}
\caption{The \textit{attack success rate vs. perturbation budget} curves of the $15$ models against dodging attacks under the $\ell_{\infty}$ norm on CFP-FP.}
\label{fig:cfp-white-d-linf-asr-pert}
\begin{center}
\includegraphics[width=0.85\linewidth]{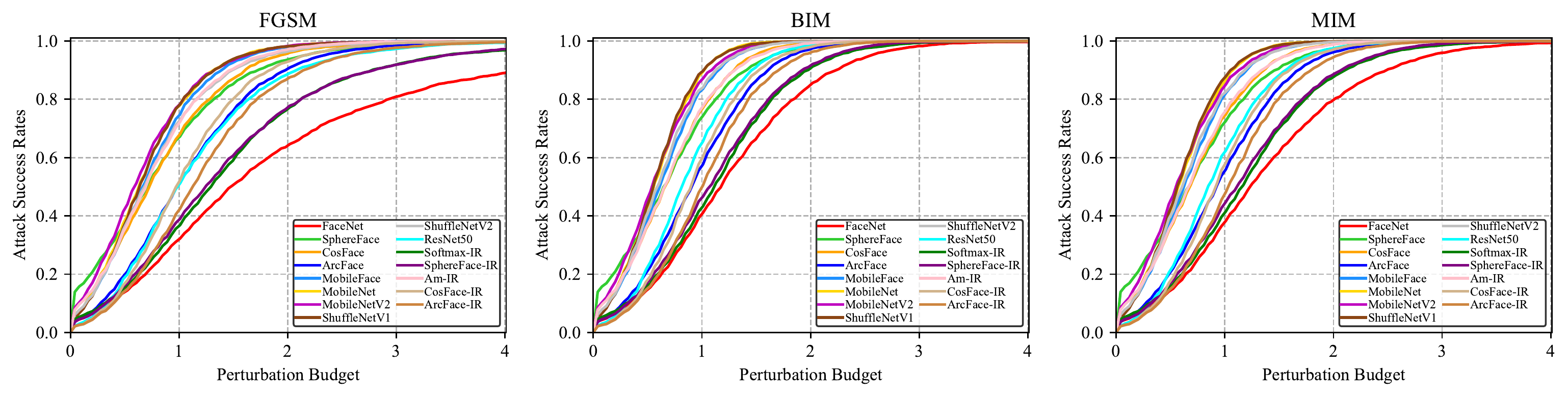}
\end{center}
\vspace{-2ex}
\caption{The \textit{attack success rate vs. perturbation budget} curves of the $15$ models against impersonation attacks under the $\ell_{\infty}$ norm on CFP-FP.}
\label{fig:cfp-white-p-linf-asr-pert}
\vspace{-2ex}
\end{figure*}

\textbf{White-box attacks:} For  CFP-FP dataset,  we also show some robustness curves for white-box attacks and provide attack success rate vs. perturbation
budget of face recognition models against dodging
and impersonation attacks under the $\ell_{2}$ and $\ell_{\infty}$ norm.   Fig.~\ref{fig:cfp-white-d-l2-asr-pert} and  Fig.~\ref{fig:cfp-white-p-l2-asr-pert} show the \emph{attack success rate vs. perturbation budget} curves of the 15 models against dodging and impersonation attacks under the $\ell_{2}$ norm. Fig.~\ref{fig:cfp-white-d-linf-asr-pert} and  Fig.~\ref{fig:cfp-white-p-linf-asr-pert} show the \emph{attack success rate vs. perturbation budget} curves of the 15 models against dodging and impersonation attacks under the $\ell_{\infty}$ norm.

\begin{figure*}[!htp]
\centering
\includegraphics[width=0.99\linewidth]{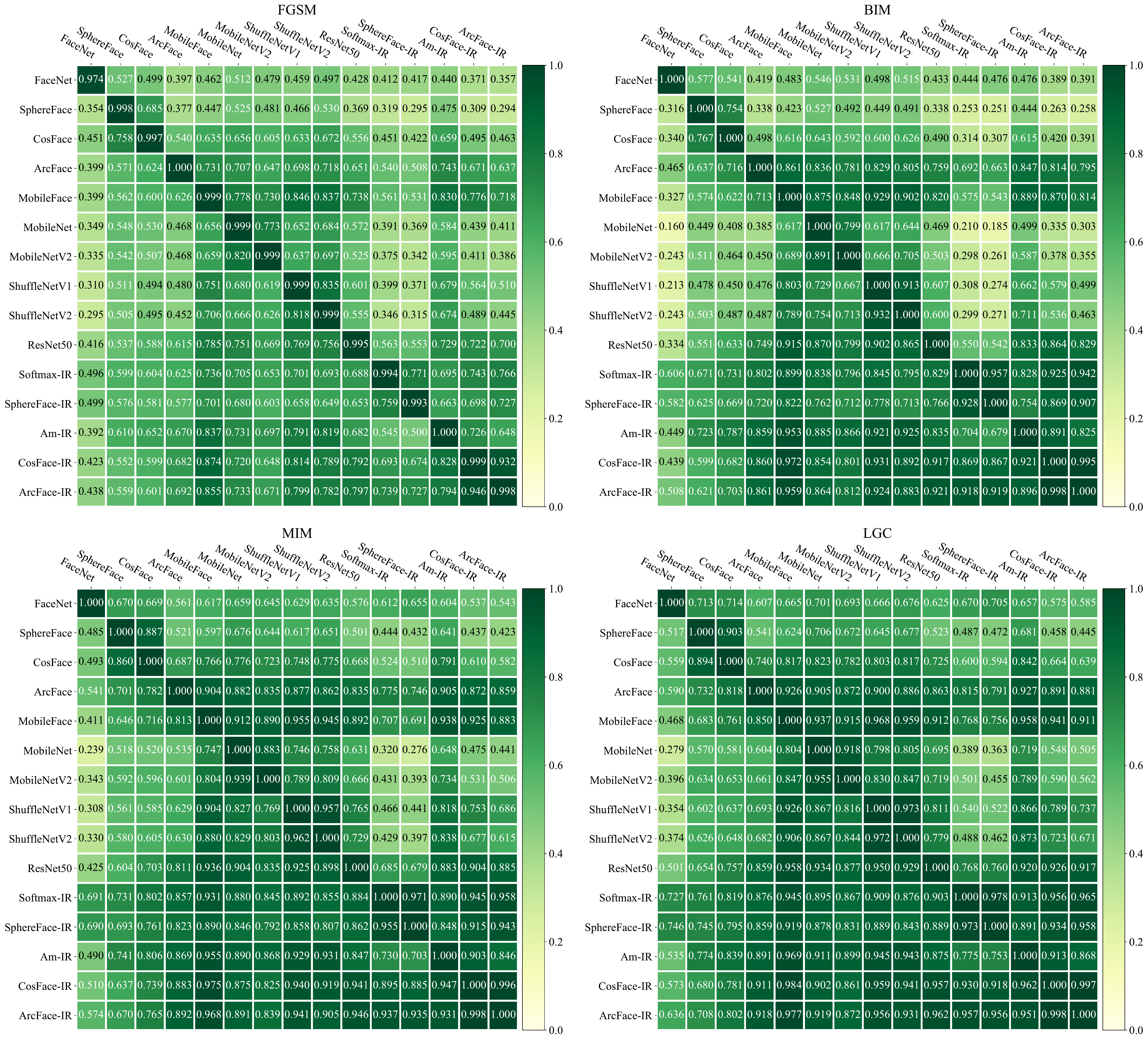}
\vspace{-2ex}
\caption{The success rates of the $15$ models against black-box dodging attacks under the $\ell_{\infty}$ norm on CFP-FP.}
\label{fig:cfp-heatmap_d}
\vspace{-1ex}
\end{figure*}

\begin{figure*}[!htp]
\centering
\includegraphics[width=0.99\linewidth]{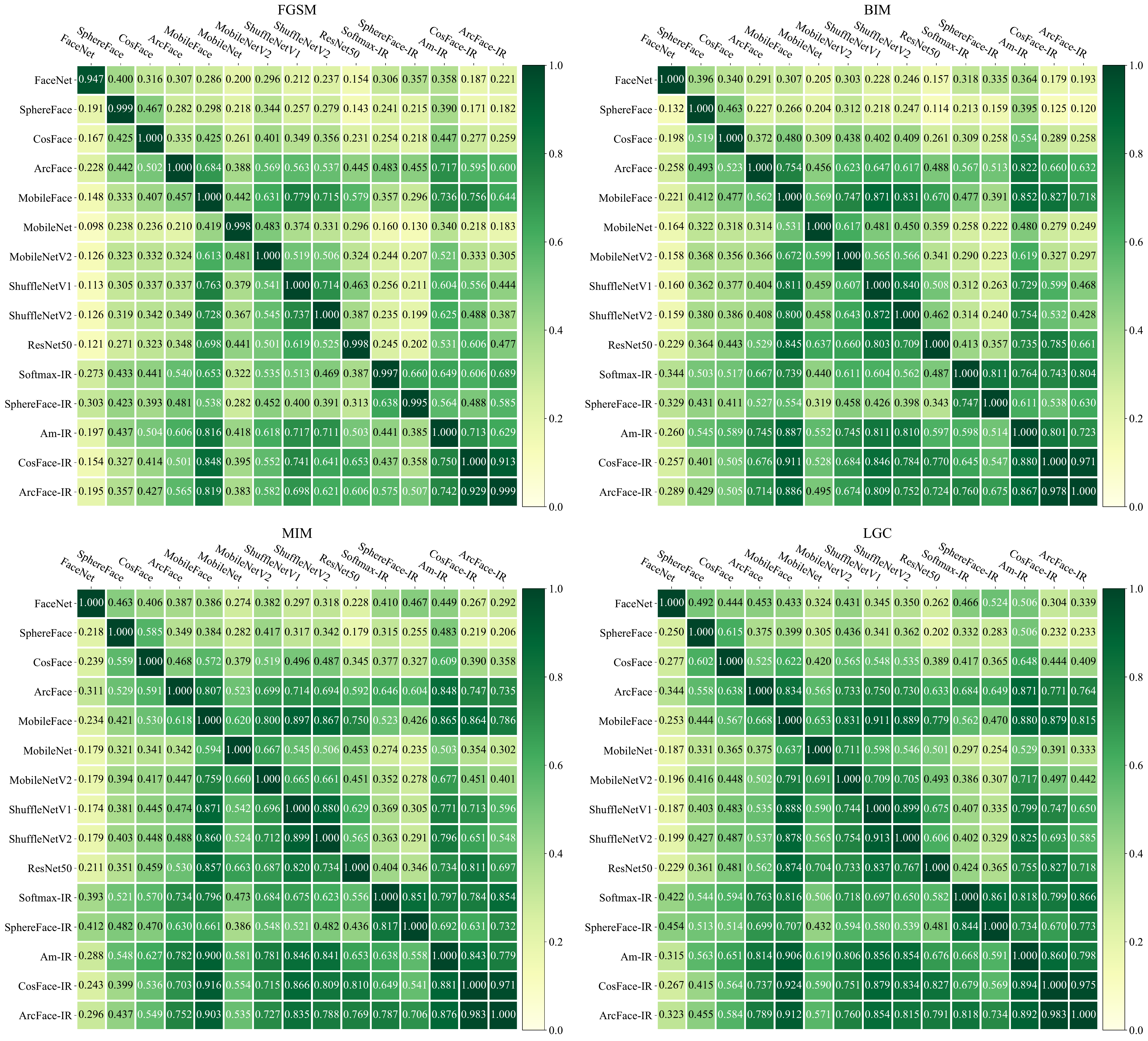}
\vspace{-2ex}
\caption{The success rates of the $15$ models against black-box impersonation attacks under the $\ell_{\infty}$ norm on CFP-FP.}
\label{fig:cfp-heatmap_p}
\end{figure*}

\textbf{Transfer-based black-box attacks:} We show the attack success rate of the 15 models against dodging
and impersonation attacks based on black-box FGSM,
BIM, MIM, and LGC methods under the $\ell_{\infty}$ norm in Fig.~\ref{fig:cfp-heatmap_d} and Fig.~\ref{fig:cfp-heatmap_p}.

\subsection{Evaluation on the MegFace dataset}

MegFace isolates pose variation with extreme poses like profile, where many features are occluded. The dataset contains $10$ frontal and $4$ profile images of $500$ individuals. Similar to LFW, the standard data protocol has defined $10$ splits, and contains $350$ pairs with the same identities and $350$ pairs with different identities.

\begin{figure*}[!htp]
\begin{center}
\includegraphics[width=0.9\linewidth]{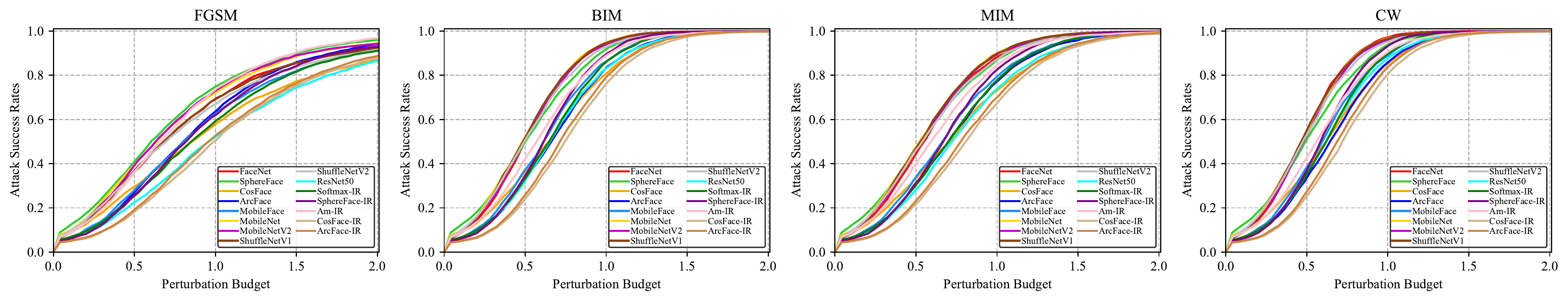}
\end{center}
\vspace{-2ex}
\caption{The \textit{attack success rate vs. perturbation budget} curves of the $15$ models against dodging attacks under the $\ell_2$ norm on MegFace.}
\label{fig:meg-white-d-l2-asr-pert}
\begin{center}
\includegraphics[width=0.9\linewidth]{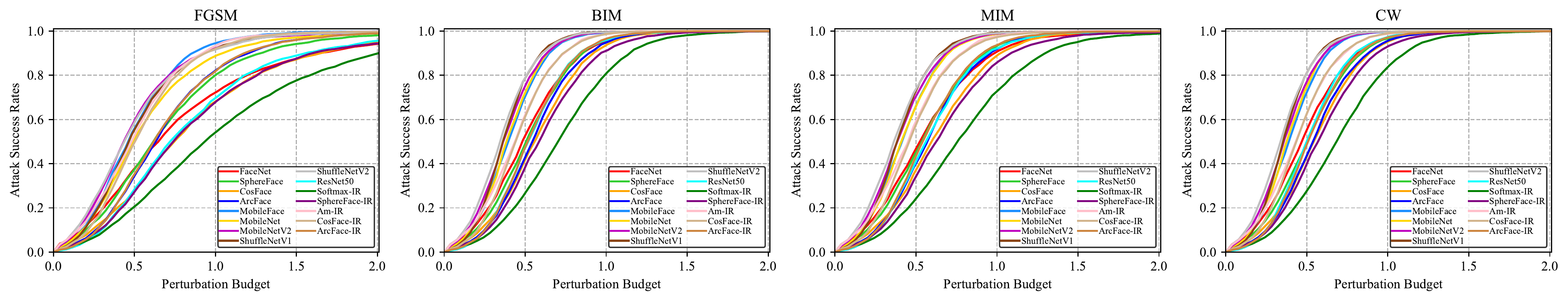}
\end{center}
\vspace{-2ex}
\caption{The \textit{attack success rate vs. perturbation budget} curves of the $15$ models against impersonation attacks under the $\ell_2$ norm on MegFace.}
\label{fig:meg-white-p-l2-asr-pert}
\vspace{-2ex}
\end{figure*}

\begin{figure*}[!htp]
\begin{center}
\includegraphics[width=0.85\linewidth]{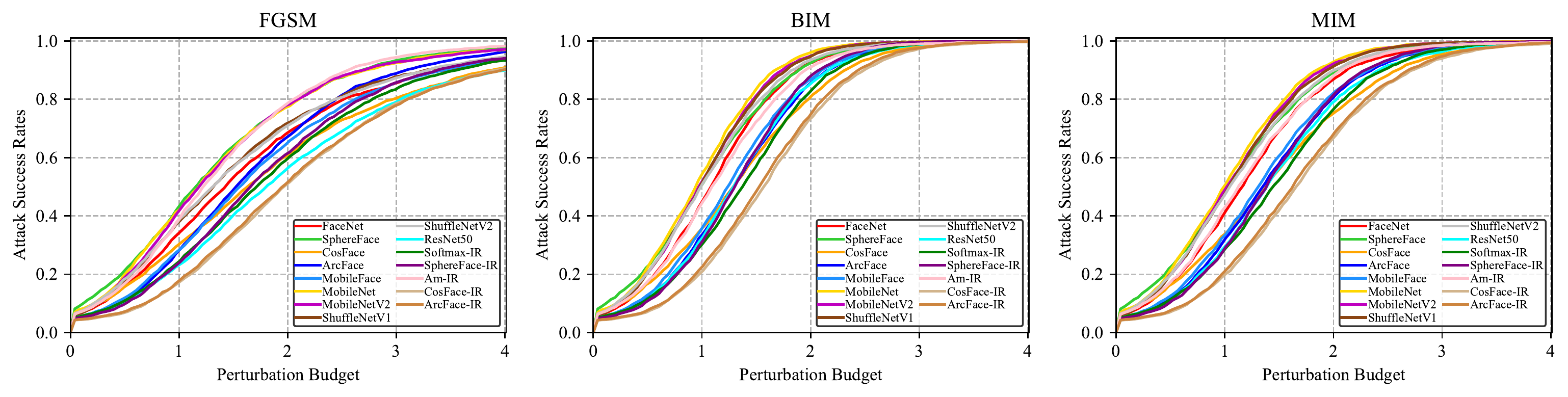}
\end{center}
\vspace{-2ex}
\caption{The \textit{attack success rate vs. perturbation budget} curves of the $15$ models against dodging attacks under the $\ell_{\infty}$ norm on MegFace.}
\label{fig:meg-white-d-linf-asr-pert}
\begin{center}
\includegraphics[width=0.85\linewidth]{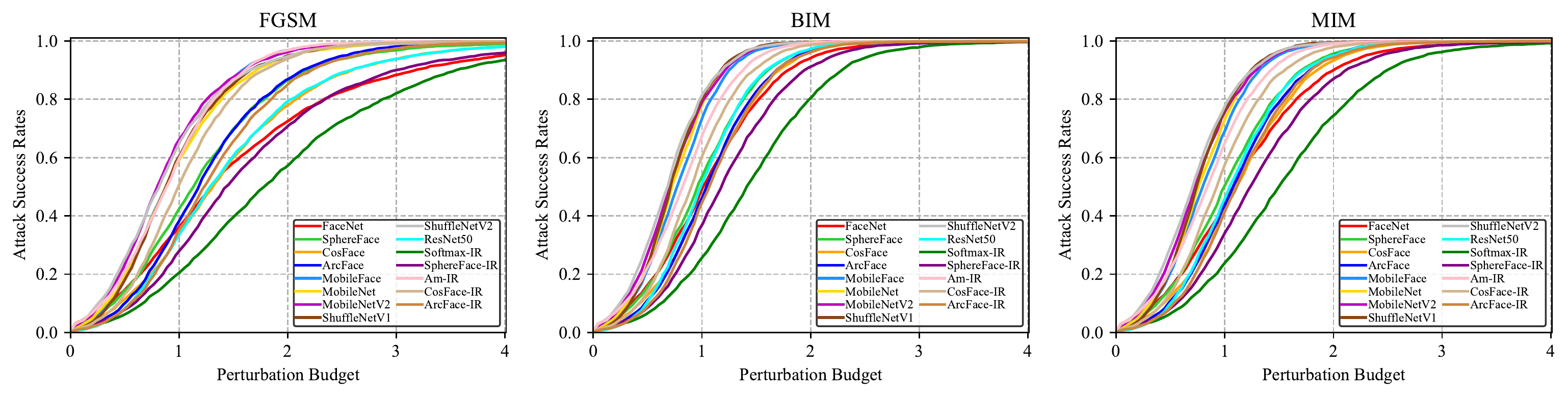}
\end{center}
\vspace{-2ex}
\caption{The \textit{attack success rate vs. perturbation budget} curves of the $15$ models against impersonation attacks under the $\ell_{\infty}$ norm on CFP-FP.}
\label{fig:meg-white-p-linf-asr-pert}
\vspace{-2ex}
\end{figure*}

\textbf{White-box attacks:} For  MegFace dataset,  we also show some robustness curves for white-box attacks and provide attack success rate vs. perturbation
budget of face recognition models against dodging
and impersonation attacks under the $\ell_{2}$ and $\ell_{\infty}$ norm.   Fig.~\ref{fig:meg-white-d-l2-asr-pert} and  Fig.~\ref{fig:meg-white-p-l2-asr-pert} show the \emph{attack success rate vs. perturbation budget} curves of the 15 models against dodging and impersonation attacks under the $\ell_{2}$ norm. Fig.~\ref{fig:meg-white-d-linf-asr-pert} and  Fig.~\ref{fig:meg-white-p-linf-asr-pert} show the \emph{attack success rate vs. perturbation budget} curves of the 15 models against dodging and impersonation attacks under the $\ell_{\infty}$ norm.

\begin{figure*}[!htp]
\centering
\includegraphics[width=0.99\linewidth]{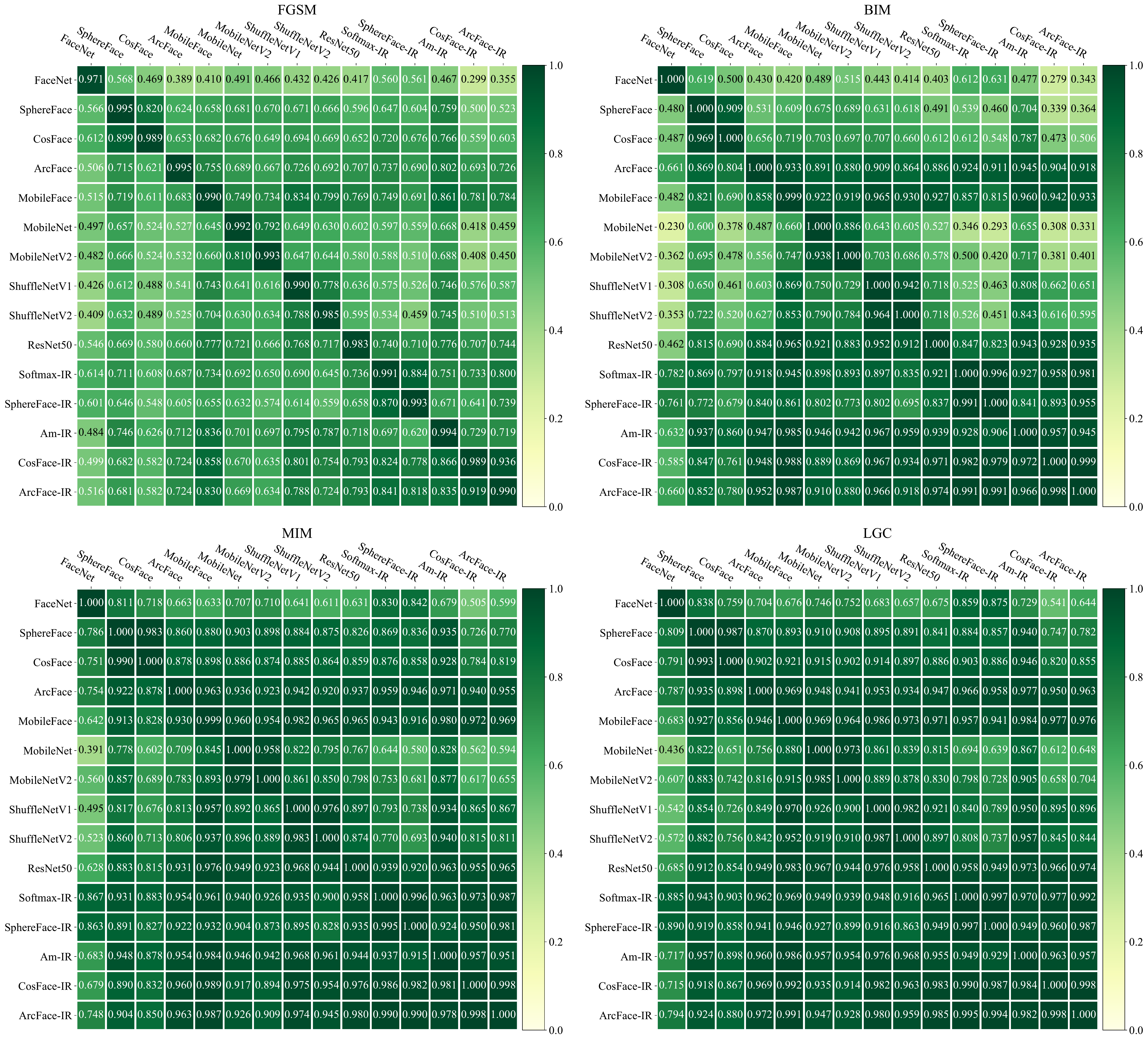}
\vspace{-2ex}
\caption{The success rates of the $15$ models against black-box dodging attacks under the $\ell_{\infty}$ norm on MegFace.}
\label{fig:meg-heatmap_d}
\vspace{-1ex}
\end{figure*}

\begin{figure*}[!htp]
\centering
\includegraphics[width=0.99\linewidth]{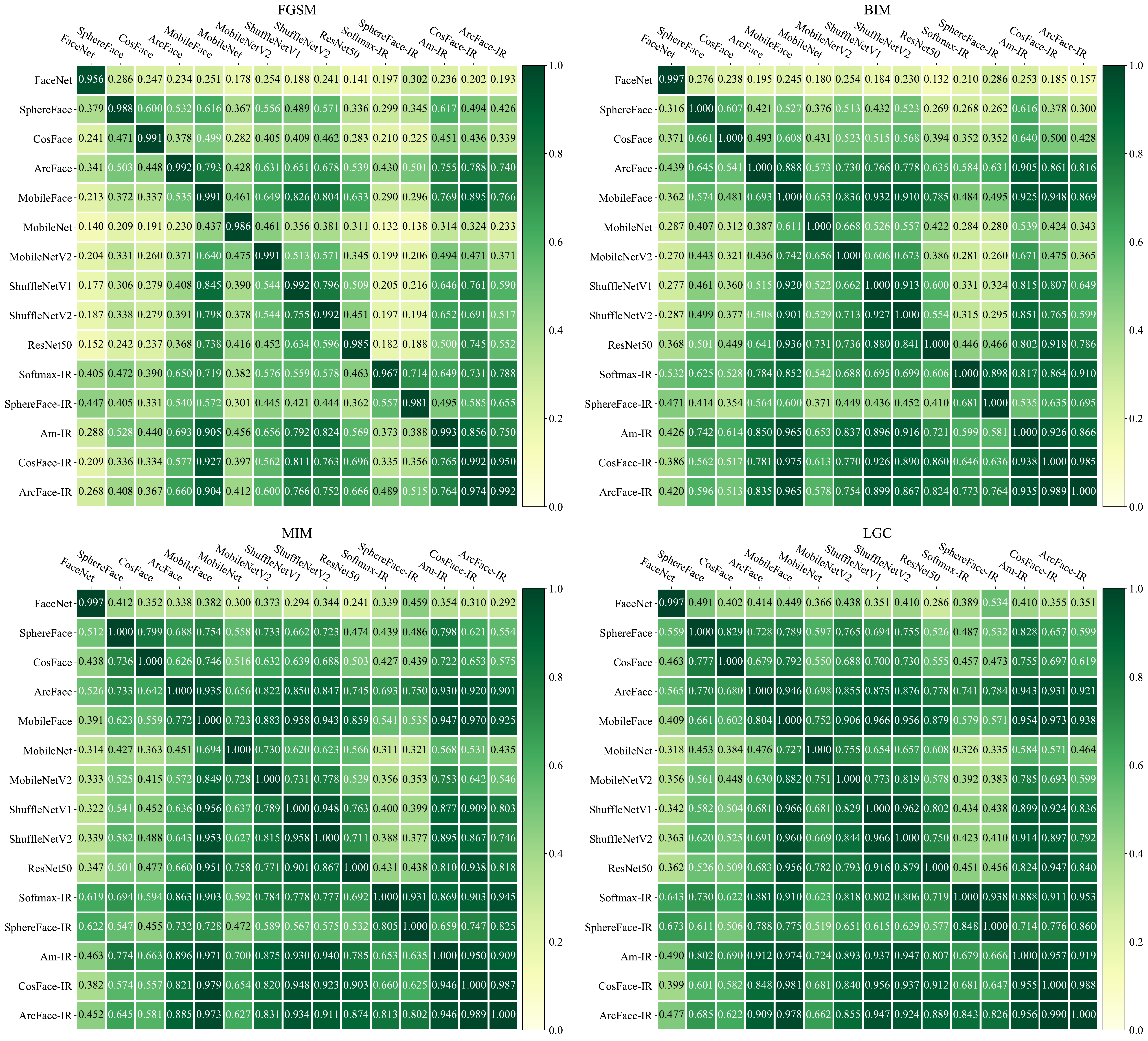}
\vspace{-2ex}
\caption{The success rates of the $15$ models against black-box impersonation attacks under the $\ell_{\infty}$ norm on MegFace.}
\label{fig:meg-heatmap_p}
\vspace{-1ex}
\end{figure*}

\textbf{Transfer-based black-box attacks:} We show the attack success rate of the 15 models against dodging
and impersonation attacks based on black-box FGSM,
BIM, MIM, and LGC methods under the $\ell_{\infty}$ norm in Fig.~\ref{fig:meg-heatmap_d} and Fig.~\ref{fig:meg-heatmap_p}.





\end{document}